\pdfoutput=1
\documentclass[10pt,twocolumn,letterpaper]{article}

\usepackage{iccv}
\usepackage{times}
\usepackage{epsfig}
\usepackage{graphicx}
\usepackage{amsmath}
\usepackage{amssymb}
\usepackage{wrapfig}

\usepackage[pagebackref=true,breaklinks=true,letterpaper=true,colorlinks,bookmarks=false]{hyperref}

\iccvfinalcopy 
\newcommand{\gitlab}{{\small\url{www.gitlab.com/haghdam/deep_active_learning}} }
\def\iccvPaperID{4697} 

\ificcvfinal\fi
\begin{document}
	
	\title{Active Learning for Deep Detection Neural Networks}
	
	\author{Hamed H. Aghdam$^1$, Abel Gonzalez-Garcia$^1$, Joost van de Weijer$^1$, Antonio M. L\'{o}pez$^{1,2}$\\
		Computer Vision Center (CVC)$^1$ and Computer Science Dpt.$^2$, Univ. Aut\`{o}noma de Barcelona (UAB)\\
		{\tt\small \{haghdam,agonzalez,joost,antonio\}@cvc.uab.es}
	}

	\maketitle
    \ificcvfinal\thispagestyle{empty}\fi
    \begin{abstract}
    \vspace*{-1mm}
		The cost of drawing object bounding boxes (i.e. labeling) for millions of images is prohibitively high. For instance, labeling pedestrians in a regular urban image could take 35 seconds on average. Active learning aims to reduce the cost of labeling by selecting only those images that are informative to improve the detection network accuracy. In this paper, we propose a method to perform active learning of object detectors based on convolutional neural networks. We propose a new image-level scoring process to rank unlabeled images for their automatic selection, which clearly outperforms classical scores. The proposed method can be applied to videos and sets of still images. In the former case, temporal selection rules can complement our scoring process. As a relevant use case, we extensively study the performance of our method on the task of pedestrian detection. Overall, the experiments show that the proposed method performs better than random selection.
	\end{abstract}
	
    \vspace*{-5mm}
 	\section{Introduction}
	\label{sec:intro}
	Having comprehensive and diverse datasets is essential for training accurate neural networks, which becomes critical in problems such as object detection since the visual appearance of objects and background vary considerably. The usual approach to create such datasets consists of collecting as many images as possible and drawing bounding boxes (labeling) for all objects of interest in all images. However, this approach has two major drawbacks. 
	
	While labeling small datasets is tractable, it becomes extremely costly when the dataset is large. For instance, according to our experiments with six labeling tools (LabelMe, VoTT, AlpsLabel, LabelImg, BoundingBox Annotation, Fast Annotation), on average, a human ({\ie} the oracle) takes a minimum of 35 seconds for labeling pedestrians of a typical urban road scene; the time can be longer depending on the tool and oracle's labeling experience.	In a dataset where hundred-thousands of images contain pedestrians, the total labeling time could be prohibitively high. 
	One way to deal with this problem is to select a \textit{random} subset for labeling. Unless the selected random subset is large, this does not guarantee that it will capture diverse visual patterns. As a result, the accuracy of the network trained on the random subset might be significantly lower than training on the full dataset.
	
	Instead of selecting the subset randomly, active learning \cite{Settles10activelearning} aims to select samples which are able to improve the knowledge of the network. To this end, an active learning method employs the current knowledge of the network to select \textit{informative} samples for labeling. The general hypothesis is that the network trained on the subset selected by active learning will be more accurate than training on a random subset of the same number of samples. This way, not only the labeling cost is reduced by selecting a smaller subset for labeling but also it guarantees that the network will be sufficiently accurate by training on this subset. 
	
\begin{figure*}
\centering
\includegraphics[width=\linewidth]{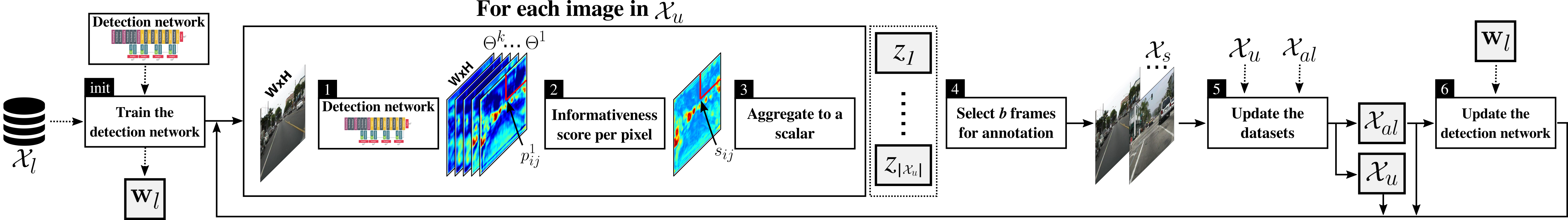}
\caption{Our method for active learning of object detectors. We start with a detector trained on a labeled dataset. Then, active learning cycles are repeated according to a preset budget. Each cycle starts by computing dense object prediction probabilities for each remaining unlabeled image. This is followed by computing pixel-level scores and aggregating them into a frame-level score. These scores are used to automatically select images for human labeling. Finally, the original detector is finetuned using the accumulated actively-labeled images.}
\label{fig:proposed_method}
\end{figure*}		

	As we will see in Section \ref{sec:related}, most work in active learning has focused on image classification. However, in general, the labeling cost is considerably higher for a detection task compared to a classification task. In this paper, we propose a method to perform active learning on detection tasks.
	\vspace{-0.3cm}
    \paragraph{Problem formulation:} We use the set of images $\mathcal{X}_l$, labeled with object bounding boxes, to train an object detector, $\Theta$, based on a convolutional neural network. 
    Afterward, we receive an unlabeled set of still images or videos $\mathcal{X}_u$. The goal is to improve the accuracy of $\Theta$ by labeling a small subset of $\mathcal{X}_u$. $\mathcal{X}_l$ and $\mathcal{X}_u$ may be from the same distribution, or there may exist a domain shift \cite{dataset_shift} between them, being the $\mathcal{X}_u$ from the domain in which $\Theta$ must perform well. In either cases, active learning aims at automatically selecting a subset $\mathcal{X}_{al}\subset\mathcal{X}_u$ such that finetuning $\Theta$ on $\mathcal{X}_{al}$ produces more accurate results than finetuning on a randomly selected subset $\mathcal{X}_{rnd}\subset\mathcal{X}_u$; where $|\mathcal{X}_{al}| = |\mathcal{X}_{rnd}|=B$ and both,  $\mathcal{X}_{al}$ and $\mathcal{X}_{rnd}$, are labeled by an oracle ({\eg} human) before finetuning. We term $B$ as \textit{total labeling budget}\footnote{There are other ways to define the labeling budget. In this paper, we use \textquotedblleft{}image-centric\textquotedblright{} definition for simplicity.}.
	
	\vspace{-0.3cm}
	\paragraph{Contribution:} In this paper, we propose a new method to perform active learning on deep detection neural networks (Section~\ref{sec:method}). In particular, given such an object detector, our method examines a set of unlabeled images to select those with more potential to increase the detection accuracy. These images are labeled and then utilized for retraining the detector. With this aim, given an image, we propose a new function to score the importance of each pixel for improving the detector. Proper aggregation of such pixel-level scores allows to obtain an image-level score. By ranking these scores, we can decide what images to select for labeling. This procedure can be performed in several iterations. Our method can be applied on both datasets of still images and videos. As a relevant use case, in Section~\ref{sec:resuls}, we carry out experiments on the task of pedestrian detection. In addition, we perform a detailed analysis to show the effectiveness of our proposed method compared to random selection and the use of other classical methods for image-level scoring. Moreover, in the case of videos, we show how our method can be easily complemented by selection rules that take into account temporal correlations. Our codes are publicly available at \gitlab. We draw our conclusion and future work in Section~\ref{sec:conclusion}.

	\section{Related Work}
	\label{sec:related}
	Most works on active learning focus on image classification.
	Gal {\etal} ~\cite{gal_al,gal_2016a,gal_droput} add a prior on the weights of image classification neural networks, sampling the weights from the dropout distribution at each evaluation. Then, the informativeness score of an unlabeled image is obtained by computing the mutual information or the variation ratio of predictions. Images are ranked according to these scores and the top $B$ are selected for labeling. The main drawback of these methods is not considering the similarity of selected samples. Therefore, they might select redundant samples for labeling. Elhamifar {\etal}~\cite{elhamifar2013convex} formulated the selection as a convex optimization problem taking into account the similarity of selected samples in the feature space as well as their informativeness score. In addition, Rohan {\etal}~\cite{icra_coreset} introduced the concept of coresets to achieve this goal. Recently, Sener and Savarese~\cite{sener2018active} cast the coreset finding problem as a k-center problem. A similar approach was utilized in~\cite{coreset_long} to do active learning over the long tail.
		
	
	Active learning was used by Lakshminarayanan {\etal}~\cite{deepmind} and Gal {\etal}~\cite{gal2017} for regression tasks, by Vondrick and Ramanan~\cite{Vondrick:2011} to select keyframes for labeling full videos with action classes, and by Heilbron {\etal} \cite{Heilbron2018} for action localization. There were different works on active learning for object detection based on hand-crafted features and shallow classifiers \cite{Abramson:2005, Sivaraman:2014}. However, to the best of our knowledge, there are only a few works on active learning for object detection based on convolution neural networks. Kao {\etal}~\cite{kao2018} rank images using the localization \textit{tightness} and \textit{stability}. The former measures how tight detected bounding boxes are, and the later estimates how stable they are in the original image and a noisy version of it. Roy {\etal}~\cite{Roy2018DeepAL} proposed \textit{black-box} and \textit{white-box} methods. Whereas the black-box methods do not depend on the underlying network architecture, white-box methods are defined based on the network architecture. Furthermore, Brust {\etal}~\cite{brust2018} computed the marginal score \cite{settles2012active} of candidate bounding boxes and integrate them using different merging functions.
	
	 There are major differences between our method and these works on active learning for deep learning based object detection. First, they incorporate commonly used score functions such as the marginal and entropy scores. In contrast, we propose a new function to compute the pixel-level score which is well suited for the task of object detection. Second, they mainly rely on simple merging functions such as average or maximum of pixel-level scores to obtain image-level scores. However, we propose another method for aggregating pixel-level scores and show its importance in our experiments. Third, when working with videos, we show how our method is well complemented with rules to avoid the selection of representative but redundant frames.

	\section{Proposed Method}
	\label{sec:method}
	Given an image $\textbf{X}$, $\textbf{x}_1 = \textbf{X}_{m_1:m_2, n_1:n_2}$ denotes a patch of it and $\textbf{x}_2 = \textbf{X}_{m_1\pm\epsilon:m_2\pm\epsilon, n_1\pm\epsilon:n_2\pm\epsilon}$ is another patch obtained by translating $\textbf{x}_1$ for $\epsilon$ pixels. We hypothesize that a detection network is likely to predict similar probability distributions for $\textbf{x}_1$ and $\textbf{x}_2$ if the appearance of these patches have been adequately seen by the network during training. Otherwise, the posterior probability distributions of $\textbf{x}_1$ and $\textbf{x}_2$ would diverge. Denoting the divergence between the posterior probabilities of $\textbf{x}_1$ and $\textbf{x}_2$ by $D(\Theta(\textbf{x}_1)||\Theta(\textbf{x}_2))$ where $\Theta()$ is the softmax output of the detection network, we assume that $D$ will be small for true-positive and true-negative predictions; while it will be high for false-positive and false-negative predictions. Thus, since our aim is to reduce the number of false-positive and false-negative predictions, we propose the active learning method illustrated in Figure \ref{fig:proposed_method} to select informative images for labeling.
	
	Initially, we assume that a labeled dataset $\mathcal{X}_l$ is used to train a network $\Theta$, giving raise to the vector of weights $\textbf{w}_l$. Active learning will start with an empty set of images $\mathcal{X}_{al}=\emptyset$ and a set of $N$ unlabeled images called $\mathcal{X}_u$. Then, active learning will proceed in cycles, where automatically selected images from $\mathcal{X}_u$ are moved to $\mathcal{X}_{al}$ after labeling. Calling $b$ the \textit{labeling budget per cycle}, and being $B$ the already introduced total budget for labeling, both expressed as number of images, we run $K=\frac{B}{b}$ active learning cycles. Note that an underlying assumption is $B \ll N$.
	
	An active learning cycle starts from Step 1 where the current unlabeled set $\mathcal{X}_u$ is processed to assign a prediction (detection) probability for each pixel of its images. Actually, it can be more than one prediction probability per-pixel for multi-resolution detection networks, which is the case we consider here; we can think in terms of matrices of prediction probabilities. In the first active learning cycle, the detection network is uniquely based on $\textbf{w}_l$. In the next cycles, these weights are modified by retraining on the accumulated set $\mathcal{X}_{al}$ of actively labeled images. Then, in Step 2, we jointly consider the spatial neighbourhood and prediction matrices to obtain a per-pixel score roughly indicating how informative may each pixel be for improving the detection network. Since we have to select full images, pixel-level scores must be converted into image-level scores. Therefore, Step 3 computes an image-level scalar score for each image $\textbf{x}_u\in\mathcal{X}_u$ by aggregating its pixel-level scores. Step 4 employs the image-level scores to select the $b$ best ranked images from $\mathcal{X}_u$ for their labeling. Denoting the set of selected $b$ images by $\mathcal{X}_s$, Step 5 sets $\mathcal{X}_{al}$ to $\mathcal{X}_{al} \cup \mathcal{X}_s$ and $\mathcal{X}_u$ to $\mathcal{X}_u - \mathcal{X}_s$. Finally, an active learning cycle ends in Step 6 after retraining $\Theta$ using $\mathcal{X}_{al}$ and $\textbf{w}_l$ as initialization weights. 
	
	Next, we explain the details of each step in our proposed method. Without loss of generality with respect to the active learning protocol, we focus on a problem which is of special relevance for us, namely pedestrian detection. We design our network and other stages of the proposed active learning method for the task of pedestrian detection, but they are extendable to multiclass detection problems. 
	
\begin{figure}[tb]
		\centering
		\includegraphics[width=0.85\linewidth]{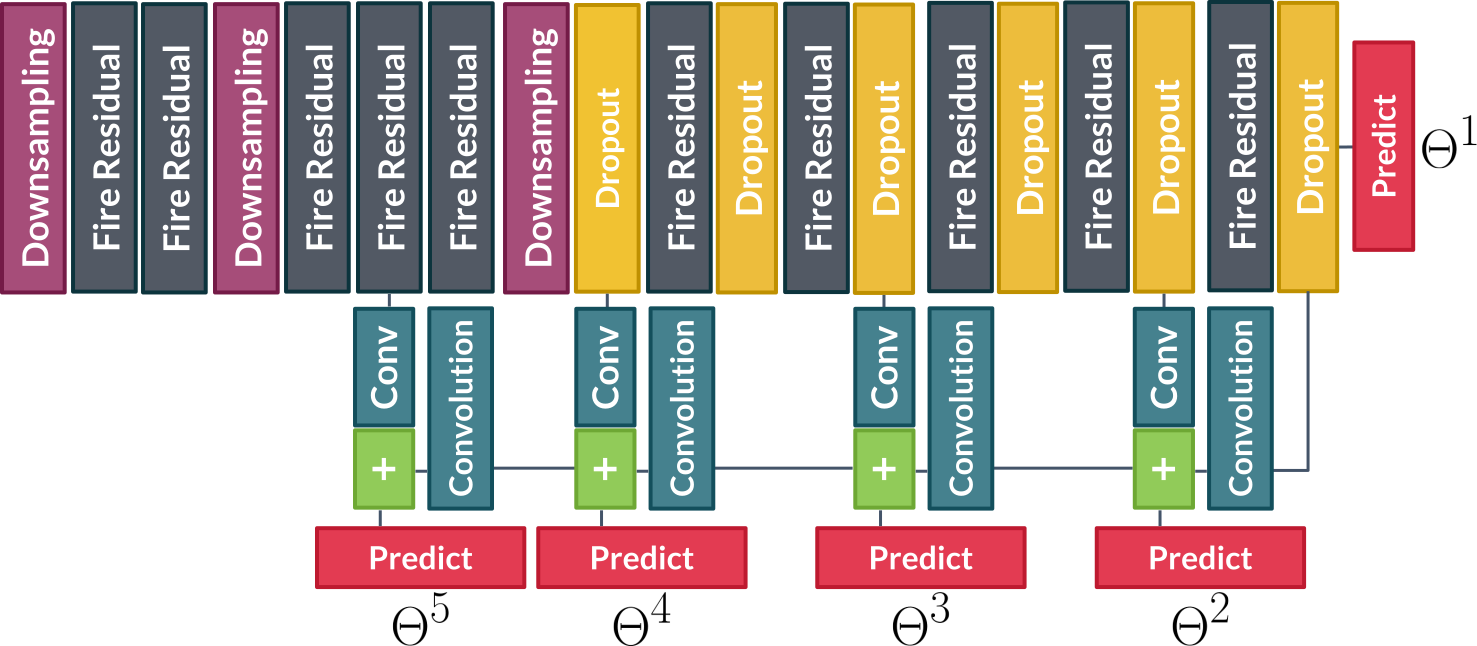}
		\caption{Overall architecture of the detection network.}
		\label{fig:network}
		\vspace{-0.3cm}
	\end{figure}

	\vspace{-0.3cm}
	\paragraph{Network architecture:} The first step in the active learning cycle computes pixel-level scores for the image. Hence, the detection network $\Theta$ must be able to compute the posterior probability for each pixel. Lin \etal \cite{fpn} proposed Feature Pyramid Networks (FPNs) with lateral connections for object detection. We utilize a similar paradigm in designing our detection network. Nonetheless, instead of using a heavy backbone network, we designed the network shown in Figure \ref{fig:network} with predictions at different levels of the decoder. Because our method requires pixel-level scores, the prediction layer must have the same size as the image. To this end, we follow \cite{deeplabv3} and resize the logits spatially using the bilinear upsampling. 
	
	Each Fire Residual module follows the same architecture indicated in \cite{squeeznet}. Also, each downsampling layer is composed of convolution and pooling layers which are applied in parallel and concatenated at the output \cite{enet}. All prediction layers \textit{share the same weights}. For this reason, there is a $1\times1$ convolution layer with 128 filters before each logits layer to unify the depth of feature maps. Interested readers can find the detail of the architecture in \gitlab. We design the network such that the prediction blocks $\{\Theta^1, \Theta^2, \Theta^3, \Theta^4, \Theta^5\}$ cover bounding boxes of size $\{270\times160, 225\times130, 145\times80, 80\times50, 55\times31\}$ respectively. Here, $\Theta^1$ indicates the prediction block connected to the last layer of the encoder and $\Theta^5$ shows the prediction block at the end of the decoder.
	
	In this paper, we have mainly focused on the task of pedestrian detection which is a binary classification problem. Thus, the depth of each logits layer is one. Also, each of them connects to a logistic loss. We do not use any bounding box regression branch in our network.
	\vspace{-0.3cm}
	\paragraph{Pixel-level scores:}
	Our goal is to select images for labeling with the highest number of false-positive and false-negative predictions. Earlier, we hypothesized that the divergence between predicted probability distributions in the neighborhood of false-positive and false-negative pixels should be high. As the result, by computing the divergence of predictions \textit{locally} we will be able to approximate the degree to which the prediction of a pixel is incorrect. 
	
	For an image of size $W\times H$, the output $\Theta^i, i=1\dots K_{\Theta}$ will be a $W\times H$ matrix of probability values, where $K_{\Theta}$ is the total number of prediction branches (matrices). For example, the element $(i,j)$ from $\Theta^3$ in our network shows how probable is that the pixel coordinate $(i,j)$ corresponds to a pedestrian that fits properly with a $145\times80$ bounding box. Given the five probability matrices, our goal is to compute the score matrix $\textbf{S}=[s_{ij}]_{{W\times H}}$ such that $s_{ij}$ shows how divergent are the predictions from each other in a local neighborhood centered at the coordinate $(i,j)$.
	
	Denoting the element $(i,j)$ of the $k^{th}$ probability matrix $\Theta^k$ with $p_{ij}^k$, the first step in obtaining the score of pixel $(m,n)$ is to compute the expected probability distribution \textit{spatially} as follows:
	\vspace{-0.2cm}
	\begin{equation}	
	\hat{p}_{mn}^k = \frac{1}{(2r+1)^2}\sum_{i=m-r}^{m+r}\sum_{j=n-r}^{n+r} p_{ij}^k.
	\vspace{-0.2cm}
	\end{equation}
	In this equation, $r$ denotes the radius of neighborhood. Next, the score of element $(m,n)$ for the $k^{th}$ probability matrix is obtained by computing
	\vspace{-0.2cm}
	\begin{equation}
	\label{eq:score_pixel_level}
	s_{mn}^k= \mathbb{H}(\hat{p}_{mn}^k) - \frac{1}{(2r+1)^2}\sum_{i=m-r}^{m+r}\sum_{j=n-r}^{n+r} \mathbb{H}(p_{ij}^k)
	\vspace{-0.2cm}
	\end{equation}
	\noindent where $\mathbb{H}$ is the entropy function. This score has been previously used by \cite{bald} and \cite{gal2017} in the task of image classification. In the case of binary classification problem, $\mathbb{H}(z)$ is defined as follows:
	\vspace{-0.2cm}
	\begin{equation}
	\mathbb{H}(z) = - z\log z - (1-z)\log(1-z).
	\vspace{-0.2cm}
	\end{equation}
	The final score for the element $(m,n)$ is obtained by summing the same element in all probability matrices.
	\vspace{-0.2cm}
	\begin{equation}
	s_{mn} = \sum_{k=1..K_{\Theta}} s_{mn}^k.
	\vspace{-0.2cm}
	\end{equation}
	Basically, the score $s_{mn}^k$ is obtained by computing the difference between the \textit{entropy of mean predictions} and the \textit{mean entropy of predictions}. As it turns out, $s_{mn}^k$ will be close to zero if the predictions are locally similar at this location. In contrary, $s_{mn}^k$ will be high if predictions deviate locally. Finally, $s_{mn}$ will be small if predictions are locally consistent in all probability matrices.
	\vspace{-0.3cm}
	\paragraph{Aggregating scores:}
	We need to rank unlabeled images based on their informativeness to select some of them in the next step. However, it is not trivial to compare two images purely using their score matrices and decide which one may provide more advantageous information to the network.
	\begin{figure}
		\centering	
		\includegraphics[width=0.6\linewidth]{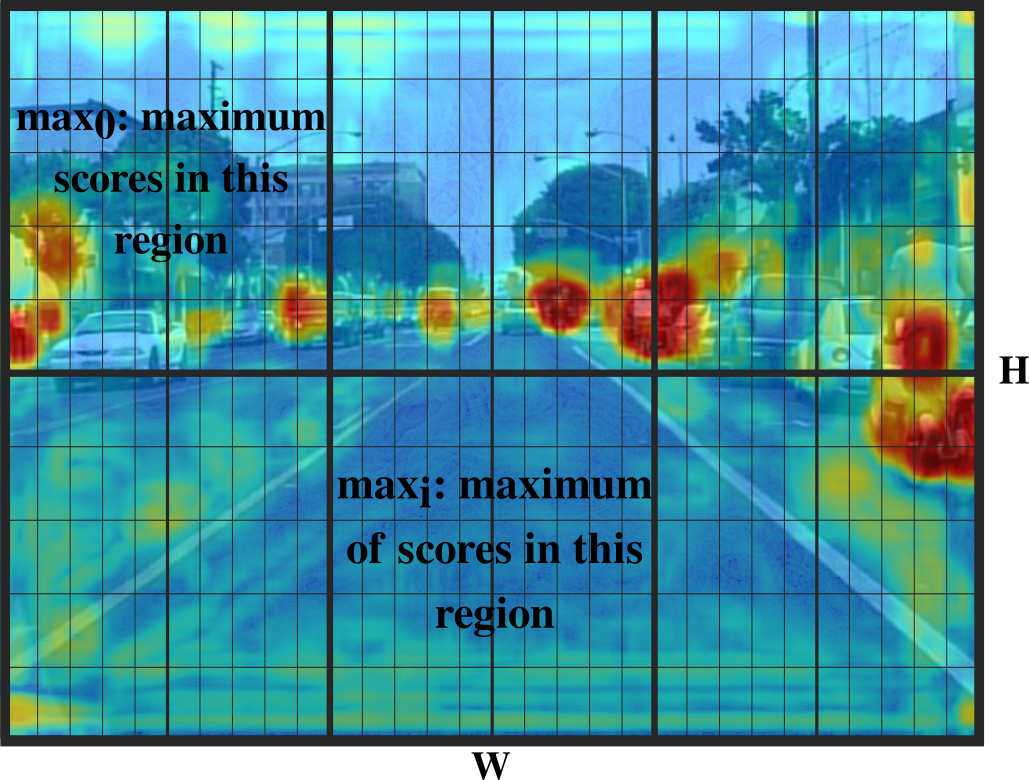}
		\caption{Aggregating pixel-level scores to an image-level score.}
		\label{fig:aggregate}
		\vspace{-0.5cm}
	\end{figure}	
	
	A straightforward solution is to aggregate pixel-level scores in the score matrix $\textbf{S}$ to a single number. To this end, we divide the score matrix $\textbf{S}$ into \textit{non-overlapping} regions. This is illustrated in Figure \ref{fig:aggregate}.	
	Then, the maximum score of each region is computed. Denoting the maximum score of the $i^{th}$ region with $s_{max}^i$, we compute the image-level score $z$ as the average of max-pooled scores:
	\vspace{-0.2cm}
	\begin{equation}
	z = \frac{1}{D_{p}} \sum_{i} s_{max}^i \enspace ,
	\vspace{-0.2cm}
	\label{eq:maxpooling}
	\end{equation}
	\noindent where $D_{p}$ is the total number of max-pooled regions.
	\vspace{-0.3cm}
	\paragraph{Selecting images:}
	As indicated in Figure \ref{fig:proposed_method}, the image-level score is computed for every sample in the unlabeled dataset $\mathcal{X}_u$. The next step is to select $b$ samples from $\mathcal{X}_u$. Here, we consider two scenarios. In the first scenario, $\mathcal{X}_u$ is composed of still images meaning that there is no temporal dependency between two consecutive samples. In the second scenario, $\mathcal{X}_u$ contains samples that are ordered chronologically. In other words, $\mathcal{X}_u$ contains video sequences.
	
	In the first scenario, top $b$ samples with the highest image-level scores are picked from $\mathcal{X}_u$. The same approach could be used for the second scenario. However, redundant samples might be selected in the second scenario if we do not incorporate temporal reasoning in the selection process. Assume that the $t^{th}$ frame in the video from $\mathcal{X}_u$ has the highest image-level score. It is likely that the $t+\triangle t$ frame has an image-level score comparable to the $t^{th}$ frame, because it is highly probable that two (or more) consecutive frames contain similar visual patterns. 
	Then, if the frames are selected without taking into account the temporal distance, this step might select many frames running on $t\pm\triangle t$ since they all may have high image-level scores. Nonetheless, only one of these frames may suffice to improve the knowledge of the network. For this reason, we add more steps for the second scenario. Specifically, we perform \emph{temporal smoothing} of the image-level scores as follows:
	\vspace{-0.2cm}
	\begin{equation}
	\hat{z}_t =\frac{1}{\sum_{i} w_i} \sum_{i=t-\triangle t}^{t+\triangle t} w_{i+\triangle t} z_i.
	\vspace{-0.2cm}
	\label{eq:ts}
	\end{equation}
	In this equation, $z_i$ denotes the image-level score of the $i^{th}$ frame, and $w_{i+\triangle t}$ shows the importance of the image-level score within a temporal window of size $2\triangle t$. In this paper, we use the Gaussian weights but other weighting functions might be also explored. Next, the top $b$ frames with the highest $\hat{z}$ are selected from $\mathcal{X}_u$ \textit{one by one} taking into account the following \emph{temporal selection rules}:
	\begin{itemize}
		\item If the $t^{th}$ frame is selected, any frame within the temporal distance $\pm \triangle t_1$ is no longer selected in the \textbf{\textit{current active learning cycle}}
		\vspace{-0.2cm}
		\item If the $t^{th}$ frame is selected, any frame within the temporal distance $\pm \triangle t_2$ is no longer selected in the \textbf{\textit{next active learning cycles}}
	\end{itemize}
	We set $\triangle t_1$ to a higher number than $\triangle t_2$. The intuition behind this heuristic is that if the $t^{th}$ frame is visually similar to the $t\pm\triangle t_1^{th}$ frame, it will adequately improve the network such that the $t\pm\triangle t_1^{th}$ will have a low image-level score in the next cycle. By setting $\triangle t_1$ to a high number, we ensure that the two frames are going to be visually different in the current cycle. On the other hand, $\triangle t_2$ sets to a small number since the $t^{th}$ and $t\pm\triangle t_2^{th}$ frames are visually almost identical. Therefore, one of them will be enough to improve the knowledge of the network in all the cycles. 
	
	More sophisticated methods such as comparing dense optical flow of two frames or image hashing might be also used to determine the similarity of two frames. Yet, it is not trivial to tell without experiments if they will work better or worse than our proposed rules. 
	\vspace{-0.3cm}	
	\paragraph{Updating the model:}
	Step 6 is to update the neural network using the currently available labeled dataset $\mathcal{X}_{al}$. In this paper, we initialize the network using the pretrained weights $\textbf{W}_l$ and train it for $T$ \textit{epochs} on $\mathcal{X}_{al}$. 
	
	\section{Experiments}
	\label{sec:resuls}
	\noindent\textbf{Datasets:} We use CityPersons~\cite{cityperson}, Caltech Pedestrian~\cite{Dollar2012PAMI} and BDD100K~\cite{bdd100} datasets. These are filtered such that only labels related to pedestrian instances are retained. Also, any pedestrian whose height is smaller than 50 pixels or its width/height ratio is not in the interval $[0.2, 0.65]$ is discarded. These choices are because of our network architecture. For a different architecture, it might be possible to ignore these filters. Table~\ref{tbl:stat_db} shows the statistics of the three datasets after applying these criteria.
	\begin{table}
		\small	
		\begin{tabular}{lccc}
			\cline{2-4}
			& CityPersons & Caltech Ped. & BDD100K\\
			\hline
			images & 1835 & 51363 &  69836\\
			instances & 7740 & 20062 & 56473 \\
			images w ped. & 1835 & 10523 & 17632 \\		
			image size & $2048\times1024$ & $640\times480$ & $1280\times720$\\
			type & Image & Video & Image\\
			\hline
		\end{tabular}	
		\caption{Statistics of the training sets.}
		\label{tbl:stat_db}
	\end{table}
	
	The CityPersons dataset is used as the initial dataset $\mathcal{X}_l$ and the Caltech Pedestrian and BDD100K are used as the \textit{unlabeled} set $\mathcal{X}_u$ during active learning cycles. As it turns out, only $20\%$ of frames in the Caltech Pedestrian dataset and $25\%$ of frames in the BDD100K dataset contain pedestrian instances. In addition, not only the size of images in $\mathcal{X}_l$ and $\mathcal{X}_u$ is different, but they are also visually distinguishable. In other words, there is a domain shift \cite{dataset_shift,Torralba2011,Xu2014DomainAO,da} between $\mathcal{X}_l$ and $\mathcal{X}_u$ that makes the active learning procedure more challenging. More importantly, while the Caltech Pedestrian dataset contains video sequences, the BDD100K is composed of still images without a clear temporal correlation between them. This will assess the effectiveness of our method on both video sequences and still images. 

	\vspace{-0.3cm}
	\paragraph{Implementation details:}	Each prediction branch in the network is connected to a sigmoid function and the network is trained by minimizing:
	\vspace{-0.2cm}
	\begin{equation}
	\small
	e(X) = \sum_{x, y \in X} \sum_{k=1}^{K_{\Theta}} -y_k\ln p_k - (1-y_k)\ln (1- p_k) + \lambda ||\textbf{W}||
	\vspace{-0.2cm}
	\end{equation}
	In this equation, $\lambda$ is the regularization coefficient, $p_k(x)=\sigma(\Theta^k(x))$ is the posterior probability and $X={(x_i, y_i)}$ is the mini-batch of training samples where $y_i\in\{0,1\}^{K_{\Theta}}$ is a binary vector. The $j^{th}$ element in this vector is 1 if the sample $x$ indicates a pedestrian that fits with the $j^{th}$ default bounding box. The above objective function is optimized using the RMSProp method with the exponential annealing rate for $T=50$ epochs. The learning rate is set to $0.001$ and it is annealed exponentially such that it reduces to $0.0001$ in the last iteration. Furthermore, the regularization term is set to $2e^{-6}$. It is important to fix a proper negative to positive (N2P) ratio ({\ie} background {\vs} pedestrians here) for the mini-batches. As can be seen in the supplementary material, N2P=15 provided the best detection accuracy in our implementation. Focusing on the active learning method, we set $r=9$ as the spatial radius of Step 2 for obtaining pixel-level scores; while we use $30\times30$ non-overlapping regions to aggregate the pixel-level scores in Step 3. Finally, since Caltech Pedestrian dataset is organized as video sequences, we set $\triangle t_1 = 15$ and $\triangle t_2=2$ to apply temporal reasoning during the frame selection (Step 4). 
	\begin{figure}
	\centering
	\includegraphics[width=0.7\linewidth]{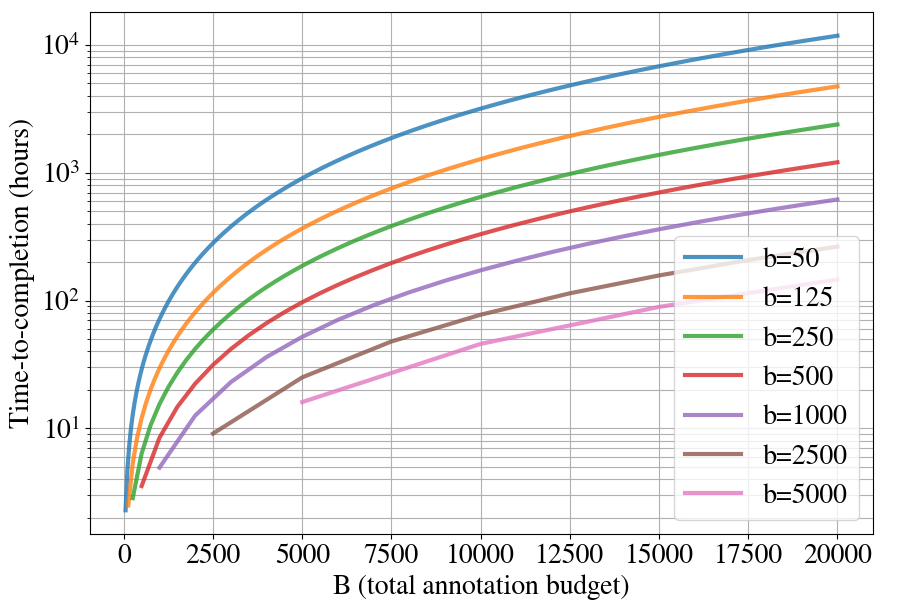}
	\caption{Time-to-completion for different $b$ and $B$.}
	\label{fig:ttc_vs_budget}
	\vspace{-0.4cm}
	\end{figure}
	
	\vspace{-0.3cm}
	\paragraph{Time-to-completion vs. budget:} Given $B$, setting $b$ is one of the important steps before performing active learning. An inappropriate $b$ value could increase the overall time-to-completion of the active learning procedure. Furthermore, setting $b$ to a high number may reduce the active learning to a sort of uniform sampling of images. 
	
	For instance, for one frame of Caltech Pedestrian dataset ($640\times480$ pixels), the time-to-completion of our network in the forward pass with double evaluation\footnote{In this paper, \textit{double evaluation} refers to evaluating the original image and its mirrored version to make predictions.} is $\sim$150 ms, and $\sim$200 ms for a forward-backward pass. Factoring out the labeling time of a frame ({\ie} by assuming that is constant), Figure \ref{fig:ttc_vs_budget} plots the overall time-to-completion of our method using different values of $b$ and $B$ for the $51,363$ frames of this dataset (Table \ref{tbl:stat_db}). Suppose $B=7500$, then, it will take $\sim$46 hours to complete the active learning procedure for $b=2500$; while it will take $\sim$1800 hours for $b=50$.
	
	On the one hand, setting $b$ to $50$ is impractical due to its high time-to-completion. Besides, adding only $50$ images at each cycle to $\mathcal{X}_{al}$ might not improve the knowledge of the network adequately. On the other hand, setting $b$ to $2500$ might reduce active learning to sampling frames uniformly (as we will explain).	Setting $b$ to $500$ is more practical since the time-to-completion is $\sim$190 hours. Moreover, adding $500$ frames to $\mathcal{X}_{al}$ at each cycle is likely to improve the accuracy of the network better. Thus, unless otherwise specified, we set $b=500$ in all our experiments.
	\begin{figure*}
		\centering
		\includegraphics[width=0.32 \linewidth]{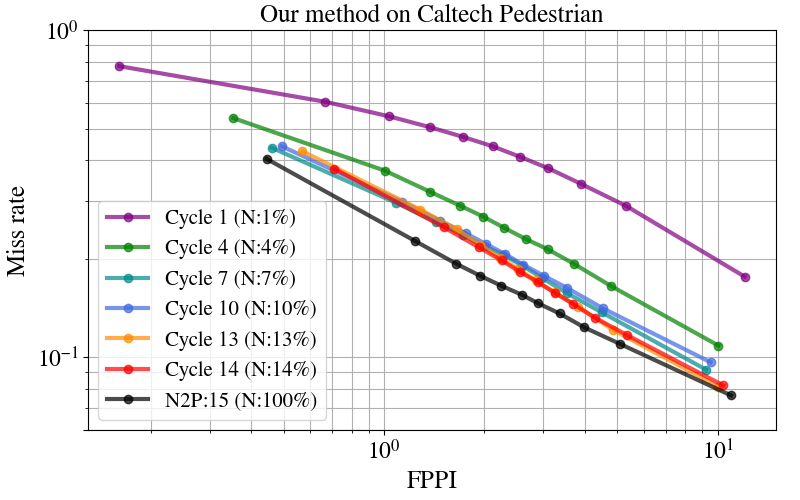}
		\includegraphics[width=0.32 \linewidth]{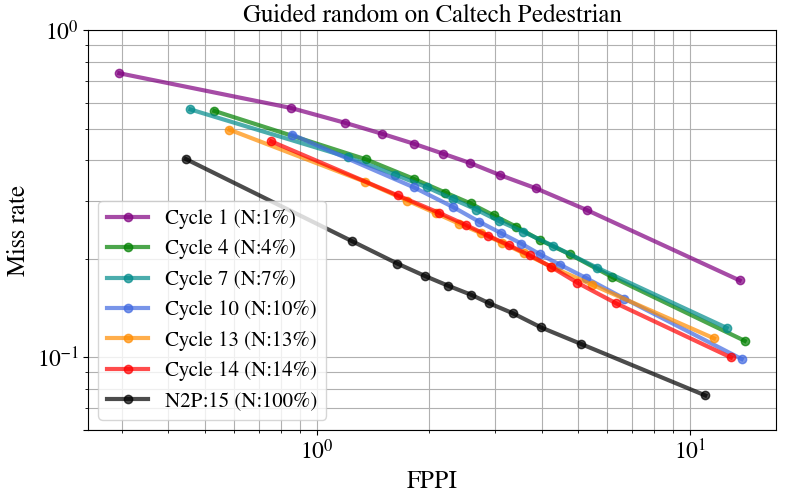}
		\includegraphics[width=0.32 \linewidth]{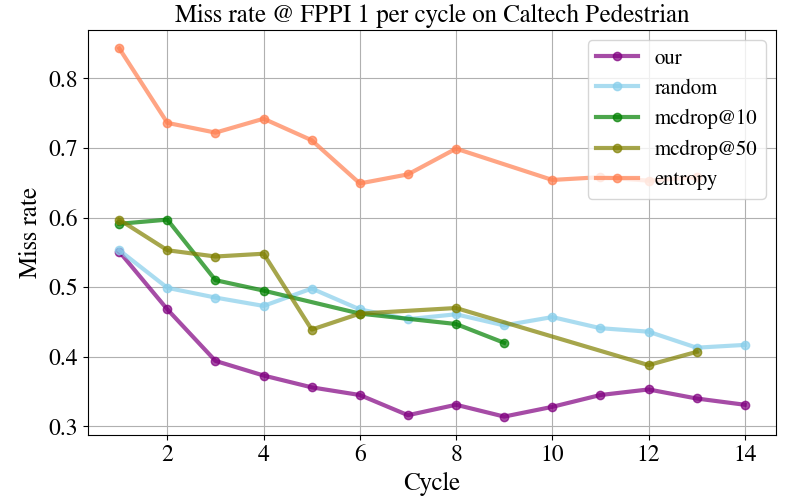}
		\caption{Performance curves of our active learning method (left) {\vs} random selection (middle), at different training cycles on the Caltech Pedestrian dataset. $N$ indicates the percentage of labeled images over the full available unlabeled training set. Thus, the black line shows the best performance that the detector can provide (it is the same in both plots). The miss rate at FPPI=1 (right) compares our active learning method with its variants based on entropy, MC-Dropout and the guided random selection.}
		\label{fig:exp_al_our_rand}	
		\vspace{-0.3cm}
	\end{figure*}

	\vspace{-0.3cm}
	\paragraph{Our method vs. random:} For comparing our method with random sampling, we assume 14 cycles. For each cycle, in the former case we apply our image selection method, while in the latter case, the selection is purely at random. In this way, we can perform per-cycle comparisons. Moreover, for a fair comparison, the same frame selection rules for videos are applied to the random selection; thus, we call it \textit{guided random}. All experiments are repeated five times. Figure \ref{fig:exp_al_our_rand} shows the mean of five runs for each method, for a selection of cycles (suppl. material includes the 14 cycles) in terms of miss rate and false-positive per image (FPPI) \cite{Dollar2012PAMI}.  	
	
	In the 1st cycle, 500 frames are selected purely using the knowledge from the CityPerson dataset. The results indicate that the frames selected by the guided random method performs comparable to the frames selected by our method. This might be due to the substantial difference between visual patterns of the CityPerson and Caltech Pedestrian datasets. In other words, the knowledge acquired from the CityPerson dataset performs similar to random knowledge in selecting informative samples at the 1st cycle. At the end of the 1st cycle, $\mathcal{X}_{al}$ contains 500 samples ({\ie} $1\%$ of the unlabeled training data) from the Caltech Pedestrian dataset. Our method exploits the knowledge obtained from current $\mathcal{X}_{al}$ to select next frames for labeling. In contrast, the guided random method does not utilize the knowledge of the network and selects the samples randomly. 
	
	At the end of the $4^{th}$ cycle, 2K frames have been selected by each of these methods. The results indicate that the $\mathcal{X}_{al}$ selected by our active learning method trains a more accurate network compared to the guided random. Finally, the results at the end of the $14^{th}$ cycle show that our method performs significantly better than the guided random on the Caltech Pedestrian dataset. This can be also seen in Figure \ref{fig:exp_al_our_rand}, showing the miss rate at FPPI=1 per cycle.
	\begin{figure*}
		\includegraphics[width=0.32\linewidth]{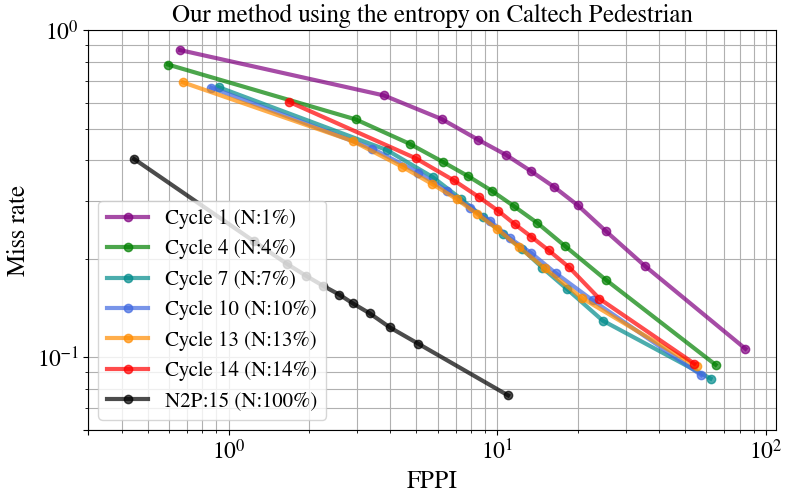}
		\includegraphics[width=0.32\linewidth]{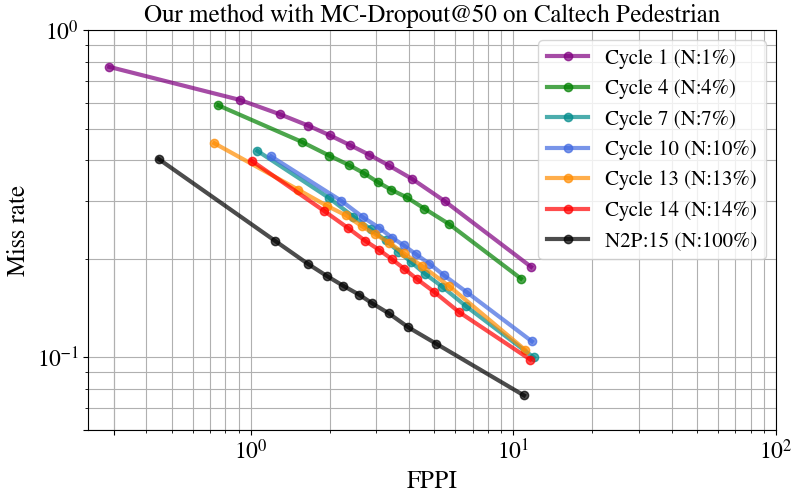}
		\includegraphics[width=0.32\linewidth]{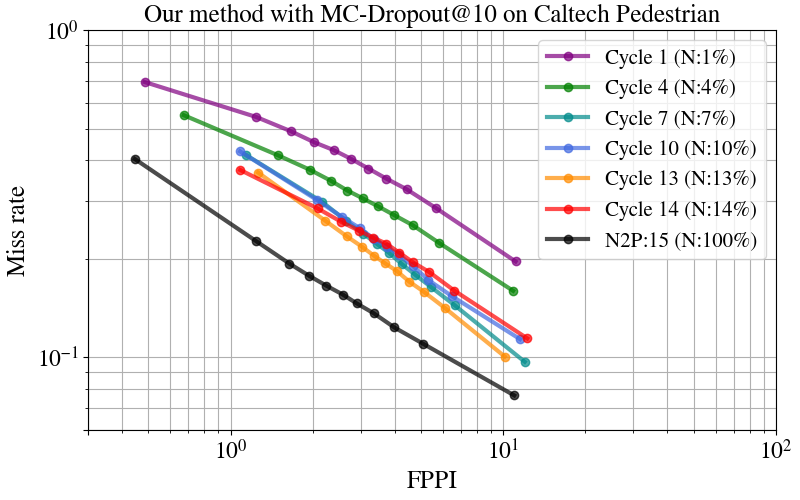}
		\caption{Performance for other score functions: binary entropy (left) and MC-Dropout with dropout ratio $50\%$ (middle) and $10\%$ (right).}
		\label{fig:caltech_ent_mc}
		\vspace{-0.5cm}
	\end{figure*}
	
	\vspace{-0.3cm}
	\paragraph{Other pixel-level score functions:} We also repeated this experiment by replacing our proposed pixel-level score function with the binary entropy and the Monte Carlo (MC) Dropout \cite{gal2017}. For the binary entropy, the pixel-level score (\ref{eq:score_pixel_level}) is replaced with $s_{mn}^k= \mathbb{H}(p_{mn}^k)$ and it is replaced with $s_{mn}^k= \mathbb{H}(\hat{p}_{mn}^k) - \frac{1}{T}\sum_{t=1}^{T} \mathbb{H}(p_{mn}^k|\textbf{w}\sim q)$
	for the MC-Dropout approach where 
	$\hat{p}_{mn}^k$ is the mean of $T$ predictions and $q$ is the dropout distribution. The main difference between this function and our proposed score function is that our function computes the divergence locally whereas MC-Dropout function computes the divergence in the same spatial location but with $T$ different predictions. We set $P=30$ and the dropout ratio to $0.5$ and $0.1$ in MC-Dropout. Figure \ref{fig:caltech_ent_mc} illustrates the results, to be compared with Figure \ref{fig:exp_al_our_rand}~(left-middle). Note how the binary entropy performs poorly even compared to guided random. MC-Dropout produces more accurate results compared to guided random but it is still less accurate than our proposed scoring function.  Figure \ref{fig:exp_al_our_rand} details more this observations comparing the respective miss rates at FPPI=1, for each cycle. 
	
\begin{figure}
		\centering
		\includegraphics[width=0.6\linewidth]{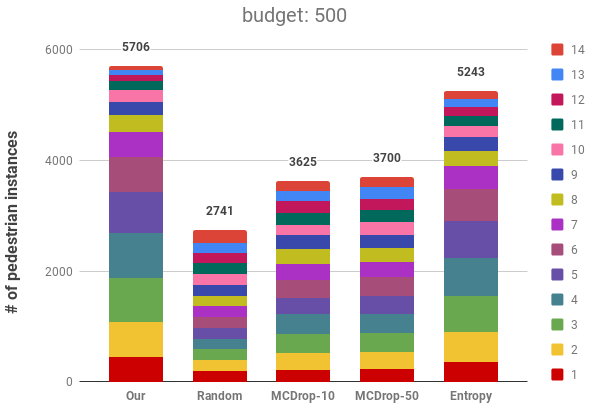}
		\caption{Number of pedestrian instances in $\mathcal{X}_s$ at each cycle.}
		\label{fig:ped_inst_our_rand}
		\vspace{-0.3cm}
	\end{figure}

	\vspace{-0.3cm}
	\paragraph{Statistics of $\mathcal{X}_{al}$:} To further analyze these methods, we computed the number of pedestrian instances selected by each method at the end of each cycle (Figure \ref{fig:ped_inst_our_rand}). 
	
	The 7K frames selected by our method contains collectively 5706 (the mean of five runs) pedestrian instances. Conversely, there are only 2741 pedestrian instances within the selected frames by the guided random method. This quantity is equal to 3700 and 5243 using our method based on the MC-Dropout and the entropy functions, respectively.
	
	Even though the method based on the entropy selects more pedestrian instances compared to guided random and MC-Dropout, it is less accurate than these two methods. This is mainly due to the fact that true-positive or true-negative candidates might have high entropy values. As the result, a frame that is processed by the network correctly might have a high image-level score and it will be selected for labeling. However, selected frames might be redundant since the network has already detected pedestrians and background correctly but with a high entropy.
	
	The intuition behind the MC-Dropout is that if the knowledge of the network about a visual pattern is precise, the predictions should not diverge if the image is evaluated several times by dropping weights randomly at each time. This is different from our method where it approximates the divergence spatially. In addition to a superior performance, our pixel-level score function is computationally more efficient than the MC-Dropout approach.	
	\begin{figure}
	\centering
	\includegraphics[width=0.66\linewidth]{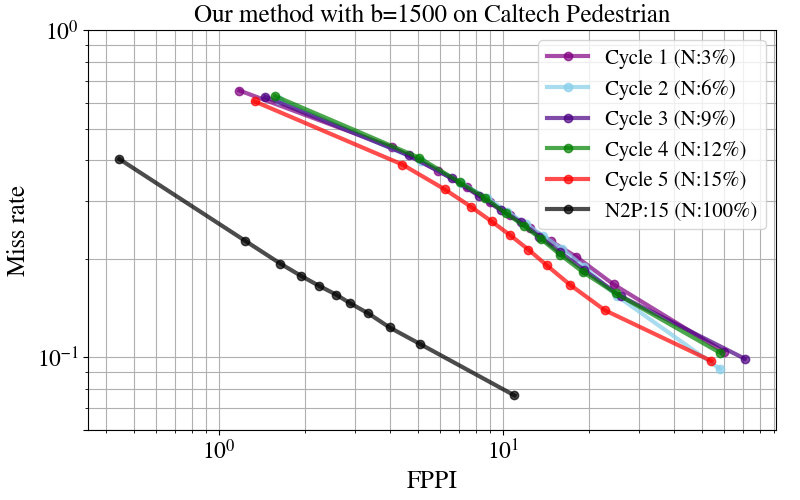}
	\caption{FPPI vs. miss rate after increasing $b$ to 1500.}
	\label{fig:exp_b1500}
	\vspace{-0.5cm}
	\end{figure}

	\vspace{-0.3cm}
	\paragraph{Importance of budget size:} Earlier in this section, we explained that setting the budget size properly is important to make the overall time-to-completion of the active learning method tractable. Here, we investigate the importance of budget size $b$ from another point of view. To this end, $b$ is increased to 1500 and the active learning method is repeated for five cycles (so labeling 1000 frames more than in previous setting). Figure \ref{fig:exp_b1500} illustrates the results. 
	\begin{figure*}
		\centering
		\includegraphics[width=0.32\linewidth]{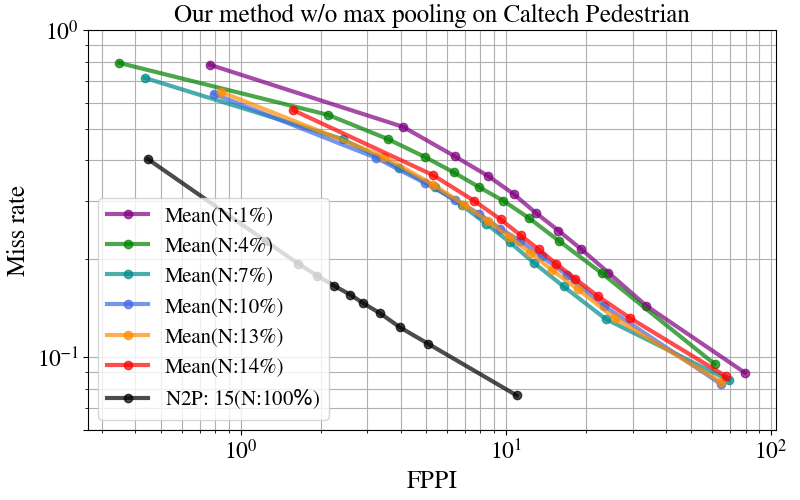}
		\includegraphics[width=0.32\linewidth]{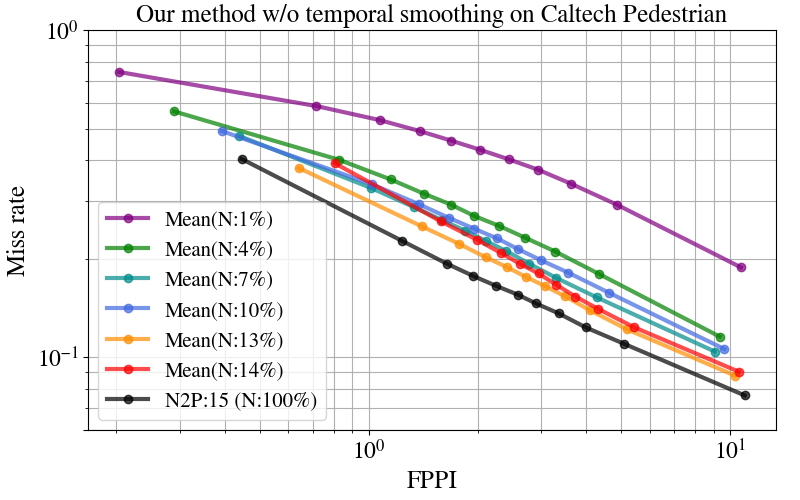}
		\includegraphics[width=0.32\linewidth]{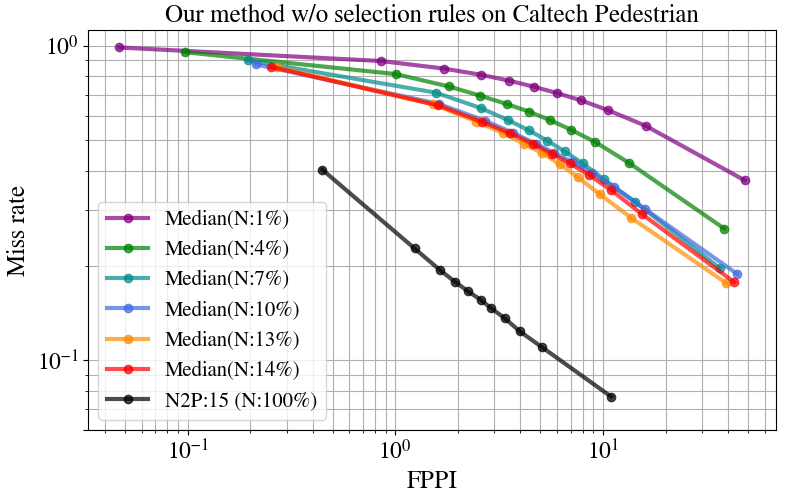}
		\caption{Performance after disabling the aggregation step (left), the temporal smoothing (middle) and the temporal selection rules (right).}
		\label{fig:exp_ablation}
		\vspace{-0.5cm}
	\end{figure*}	
	\begin{figure}
		\centering
		\includegraphics[width=0.7 \linewidth]{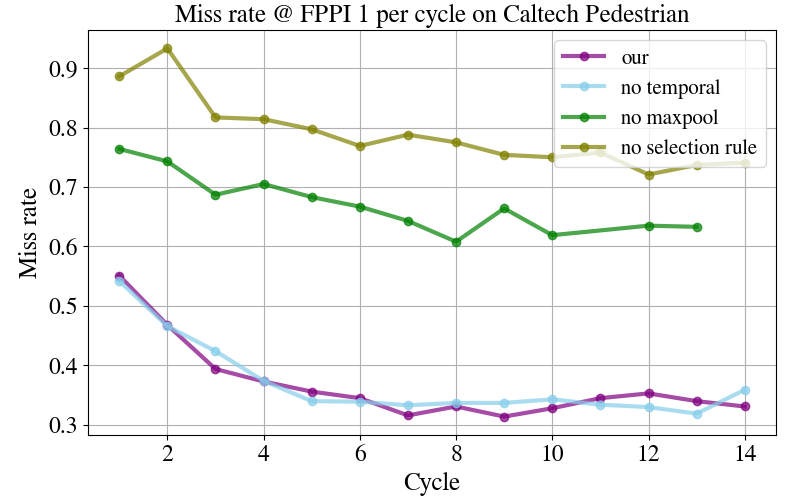}\\
		\caption{Miss rate at FPPI 1, for each cycle, when changing/disabling intermediate rules of our method.}
		\label{fig:exp_al_our_vs_ablation_fppi1}	
		\vspace{-0.3cm}
	\end{figure}

	At the $5^{th}$ cycle, 7500 frames are selected for labeling. Nevertheless, the network trained on the $\mathcal{X}_{al}$ selected by our method with $b=1500$ is less accurate than the network trained on $\mathcal{X}_{al}$ selected by the guided random method with $b=500$. This is mainly due the first criteria in the selection rules and the redundancy in $\mathcal{X}_s$ in the current cycle. 
	
	According to the first criteria, once a frame is selected, any frame within the temporal distance $\pm15$ from the selected frame will be skipped in the current cycle. When the budget size $b$ is high, this forces the selection process to perform similar to sampling frames uniformly. Furthermore, the knowledge of the network is superficial at the first cycle and it might not be able to estimate the informativeness of each frame properly. When $b$ is high, some of the selected frames in the first cycles might be redundant. Yet, by setting $b$ to a smaller value, the algorithm is able to select the adequate amount of frames to improve its knowledge and reduce the chance of selecting redundant samples.

	\vspace{-0.3cm}
	\paragraph{Ablation study:} Next, we study the importance of each step in our proposed method. In each experiment, one step is disabled while the others remain active. First, the aggregation step (max-pooling, Eq. \ref{eq:maxpooling}) for computing the image-level score was changed to just averaging the pixel-level scores. Second, the temporal smoothing step (Eq. \ref{eq:ts}) was not applied. Third, the temporal selection rules were not applied. Figures \ref{fig:exp_ablation} and \ref{fig:exp_al_our_vs_ablation_fppi1} illustrate the results. We see that temporal smoothing does not seem to help considerably, while max-pooling based aggregation and temporal selection rules are critical in the Caltech Pedestrian dataset.
	
	\vspace{-0.3cm}
	\paragraph{BDD100K dataset:} We also applied our method on the BDD100K dataset which contains only still images. Thus, temporal smoothing and temporal selection rules are not applied. We first trained our network on this dataset in order to estimate the lower bound error for the active learning method. Results for different N2P values are illustrated in the supplementary material, N2P=15 was again an optimum. 
    However, the accuracy of our detection network drops on this dataset compared to the Caltech Pedestrian dataset. This is mainly due to the fact that BDD100K is visually more challenging, thus, we argue that our current network is not adequately expressive to learn complex mappings and provide a good accuracy. 
	\begin{figure}
		\centering
		\includegraphics[width=0.7\linewidth]{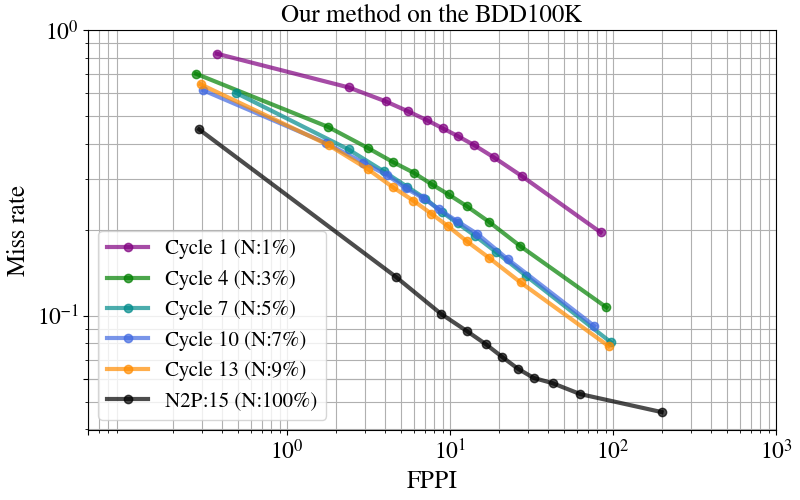}\\
		\includegraphics[width=0.7\linewidth]{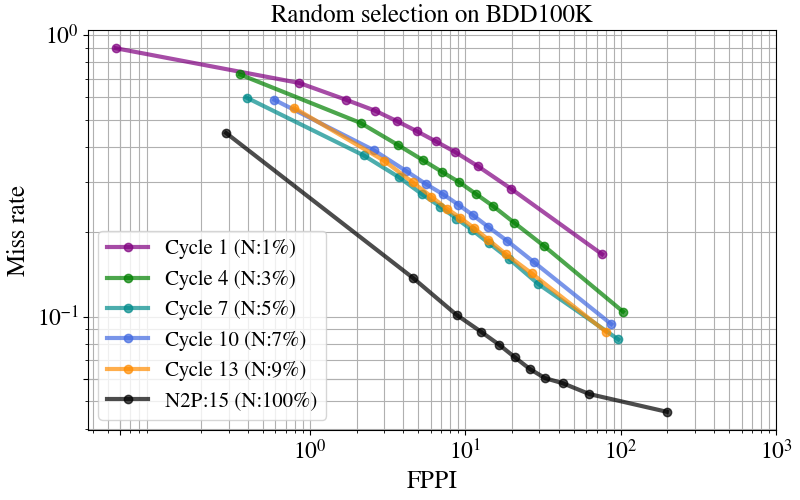}
		\caption{Performance of our method (top) and random selection (bottom) at different cycles on the BDD100K dataset.}
		\label{fig:exp_bdd_n2p}
		\vspace{-0.5cm}
	\end{figure}
	\begin{figure}
		\centering
		\includegraphics[width=0.7\linewidth]{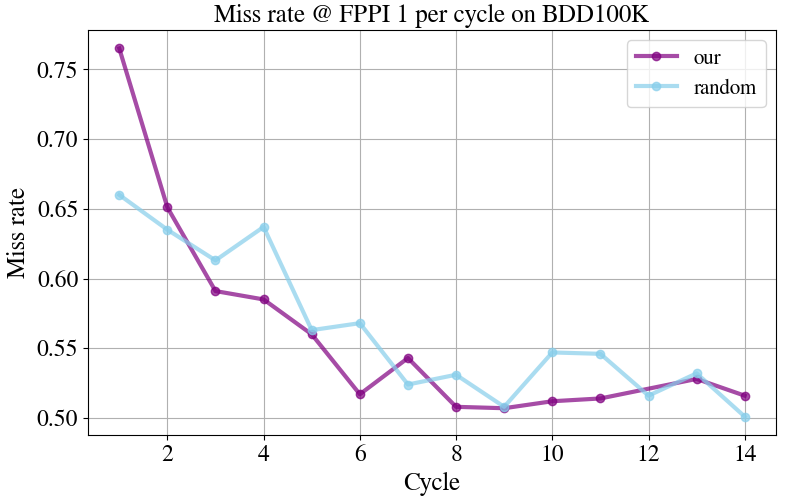}
		\caption{Miss rate FPPI=1, per cycle, for our method and random selection.}
		\label{fig:exp_bdd_fppi}
		\vspace{-0.5cm}
	\end{figure}

	Figures \ref{fig:exp_bdd_n2p} and \ref{fig:exp_bdd_fppi} illustrate the accuracy of our active learning method and random selection, including the accuracy of our network when using the full available labeled training set and N2P=15. Random selection performs better than our method at the end of the 1st cycle. However, our method performs slightly better starting from the 2nd cycle. In contrast to the Caltech Pedestrian dataset case, the improvement of our method over random is just slight.
		
	One reason might be due to the network architecture which has a high bias on the BDD100K dataset. When the bias is high, the majority of visual patterns will be informative to the network. Nevertheless, the network will not be able to learn more complex mappings from new samples. Consequently, visual patterns similar to samples in $\mathcal{X}_{al}$ will still have high scores in next cycles. In other words, redundant samples are likely to be selected in the next cycle if the network has a high bias. Overall, to solve this problem we must start by designing a more accurate network. Thus, we plan to consider a network with higher capacity, since otherwise we think it will be very difficult for any active learning method to reach the accuracy of using the 100\% of the labeled data without significantly increasing the number of cycles (in Cycle 14 we only use the 13\% of data here). 
	
	
	\section{Conclusion}
	\label{sec:conclusion}
	We have proposed an active learning method for object detectors based on convolutional neural networks. Overall, it outperforms random selection provided that the detector has sufficient capacity to perform well in the targeted domain. Our method can work with unlabeled sets of still images or videos. In the latter case, temporal reasoning can be incorporated as a complementary selection. We have performed an ablative study of the different components of our method. We have seen that specially relevant is the proposed max-pooling based aggregation step, which outperforms other proposals in the literature. As a relevant use case, our experiments have been performed on pedestrian detection facing domain shift alongside. In fact, our method can be generalized to segmentation problems as well as multi-class object detection and this is what we consider as our immediate future work.

	\begin{small}
	\noindent\textbf{Acknowledgements.} The authors would like to sincerely thank Audi Electronics Venture GmbH for their support during the development of this work. As CVC members, the authors also thank the Generalitat de Catalunya CERCA Program and its ACCIO agency. Antonio acknowledges the financial support by the Spanish project TIN2017-88709-R (MINECO/AEI/FEDER, UE) and Joost the project TIN2016-79717-R. Antonio thanks the financial support by ICREA under the ICREA Academia Program. 
	\end{small}
	
	\bibliographystyle{ieee_fullname}
	\bibliography{all.bib}
\end{document}


	\title{Active Learning for Deep Detection Neural Networks \\ (Supplementary Materials)}
	
	\author{Hamed H. Aghdam$^1$, Abel Gonzalez-Garcia$^1$, Joost van de Weijer$^1$, Antonio M. L\'{o}pez$^{1,2}$\\
		Computer Vision Center (CVC)$^1$ and Computer Science Dpt.$^2$, Univ. Aut\`{o}noma de Barcelona (UAB)\\
		{\tt\small \{haghdam,agonzalez,joost,antonio\}@cvc.uab.es}
	}

\maketitle



\section{Caltech Pedestrian dataset}

\paragraph{Lower-bound error.}
$\mathcal{X}_u$ is the set of unlabeled images that must be partially labeled by following our active learning method. In our experiments,  $\mathcal{X}_u$ corresponds to either Caltech Pedestian dataset or BDD100K. This section focuses on the former, next section in the later. 

\begin{figure}[t!]
		\centering
		\includegraphics[width=0.99\linewidth]{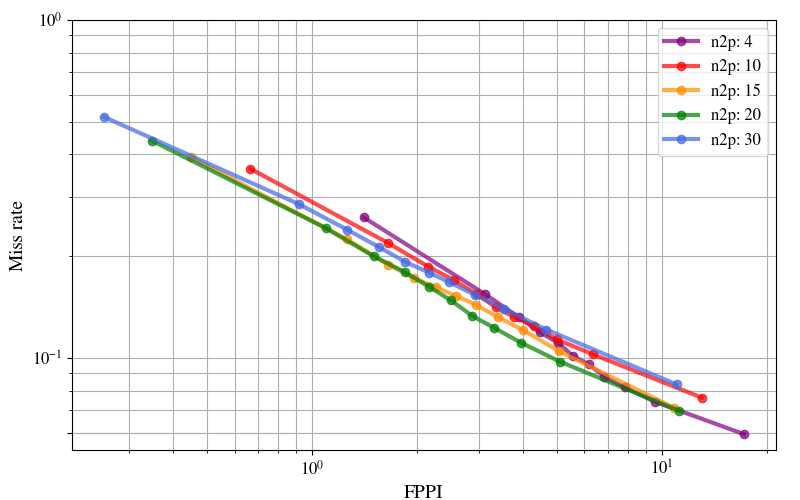}
		\caption{Pedestrian detection performance of our base neural network on the Caltech Pedestrian dataset, using different N2Ps (negative to positive ratios).}
		\label{fig:exp_al_our_rand}	
	\end{figure}

In order to estimate a lower-bound error for our active learning method, we trained our detection network on the Caltech Pedestrian dataset using all labeled training frames and evaluated on its test set. Specifically, the network is trained using different negative-to-positive (N2P) ratios. For each N2P, we trained the network three times. Figure \ref{fig:exp_al_our_rand} illustrates the mean false positive per image (FPPI) {\vs} the miss rate \cite{Dollar2012PAMI}.

	\begin{figure}[t!]
    \centering
    \includegraphics[width=0.99\linewidth]{images/ped_instance_our_vs_random.png}\\
    \includegraphics[width=1\linewidth]{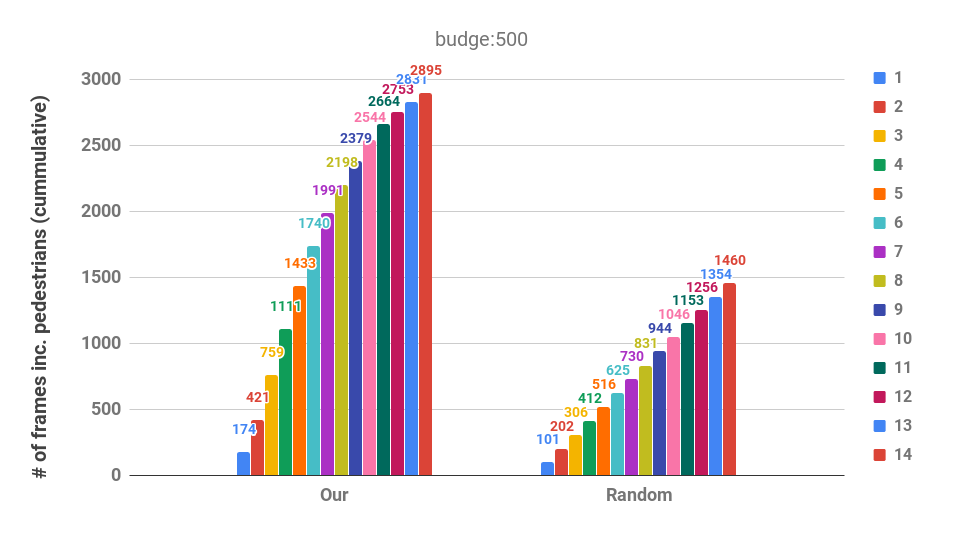}
    \caption{Statistics of $\mathcal{X}_{al}$ at each cycle in terms of number of pedestrian instances (top) and  number of images containing at least one pedestrian instance (bottom) when $\mathcal{X}_u$ is the Caltech Pedestrian dataset. Bars correspond to cycles.}
    \label{fig:caltech_stat_frames}
\end{figure}
	
We observe that the N2P affects the overall performance of the network. The minimum FPPI is greater than one when the N2P is fixed to 4. Moreover, as the N2P increases, the minimum FPPI is reduced. However, the maximum FPPIs are comparable when the N2P is greater than 10. Another way to compare these curves is to study their miss rates at $FPPI=10^0$. This way, the network trained using N2P=15 produces the best results (lower miss rate) \footnote{The high value for N2P also depends on our implementation which is available at \gitlab}.

\begin{figure*}
    \centering
    \includegraphics[width=0.32\linewidth]{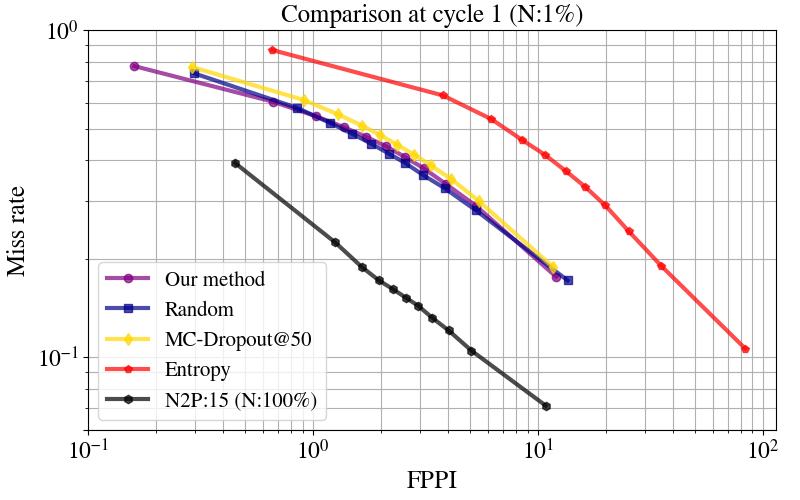}
    \includegraphics[width=0.32\linewidth]{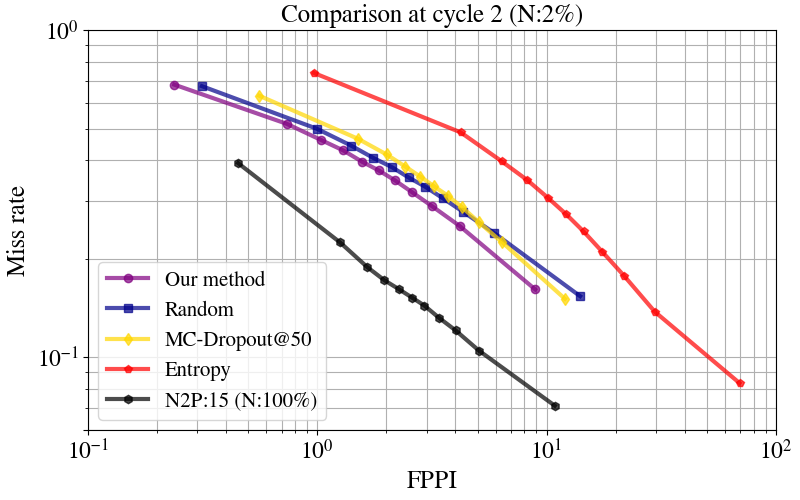}
    \includegraphics[width=0.32\linewidth]{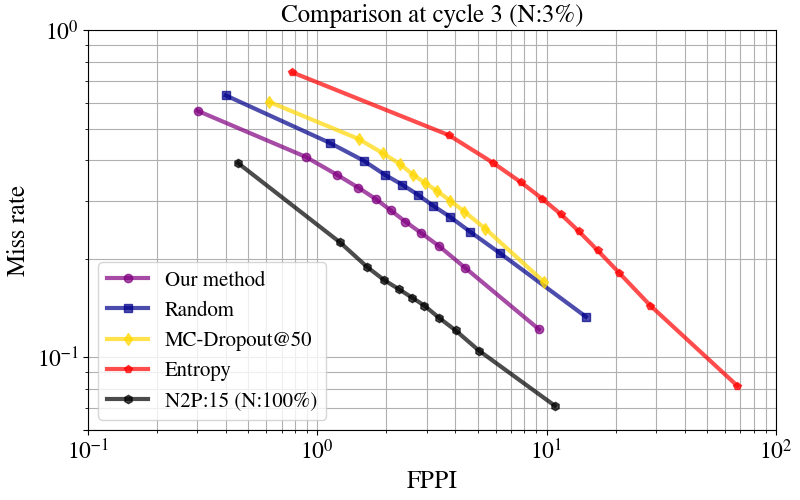}\\
    \includegraphics[width=0.32\linewidth]{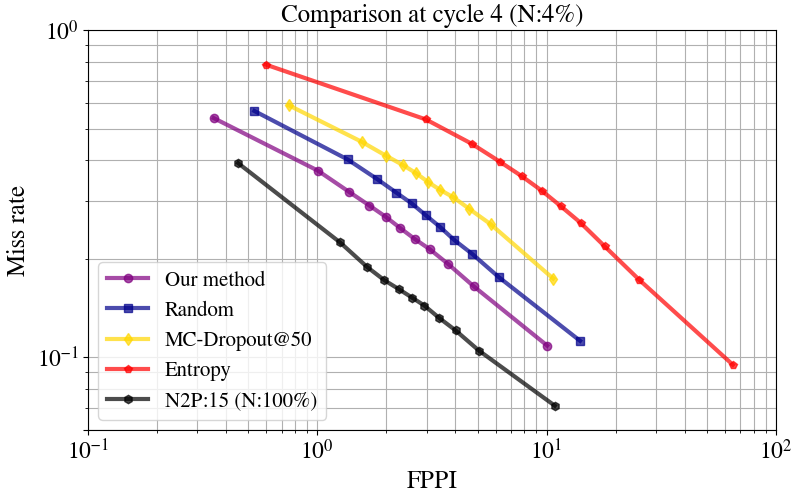}
    \includegraphics[width=0.32\linewidth]{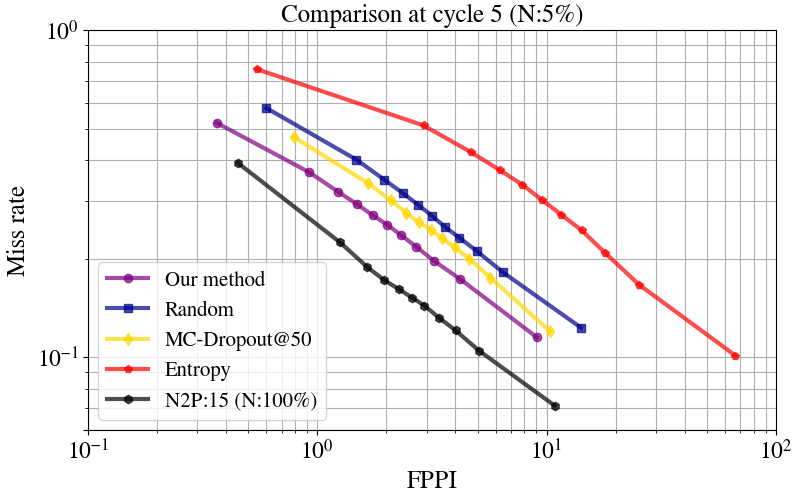}
    \includegraphics[width=0.32\linewidth]{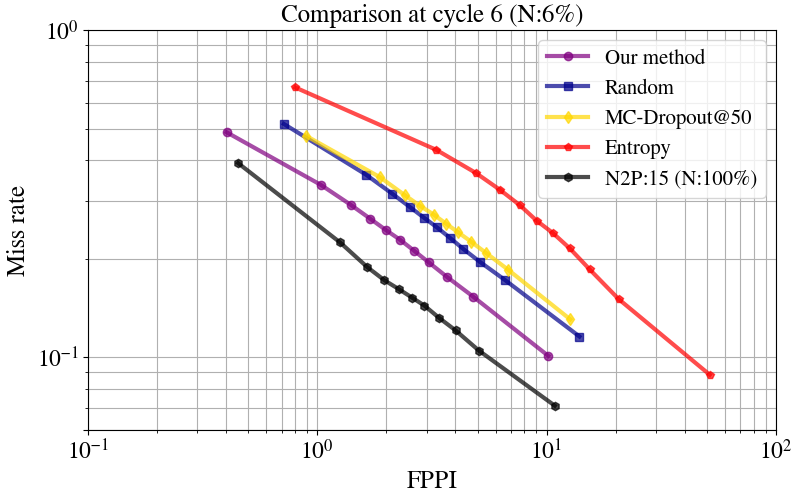}\\
    \includegraphics[width=0.32\linewidth]{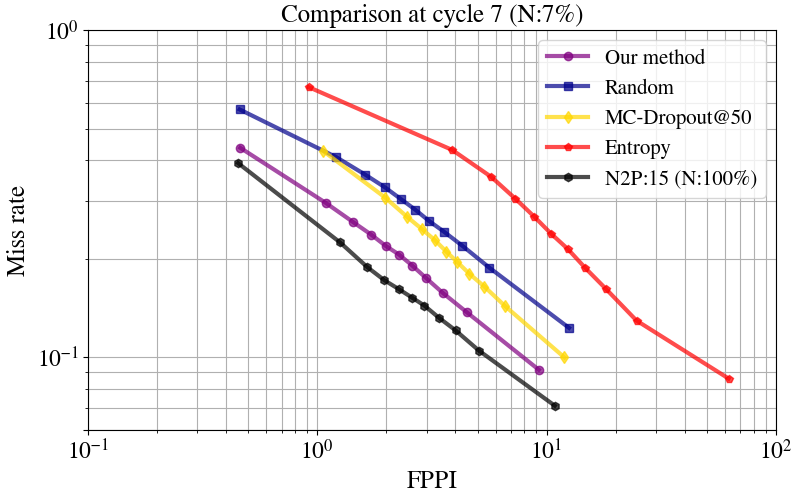}
    \includegraphics[width=0.32\linewidth]{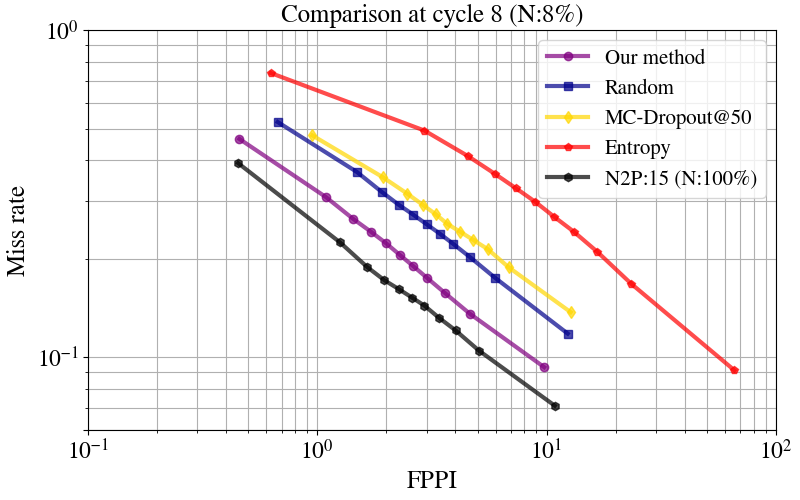}
    \includegraphics[width=0.32\linewidth]{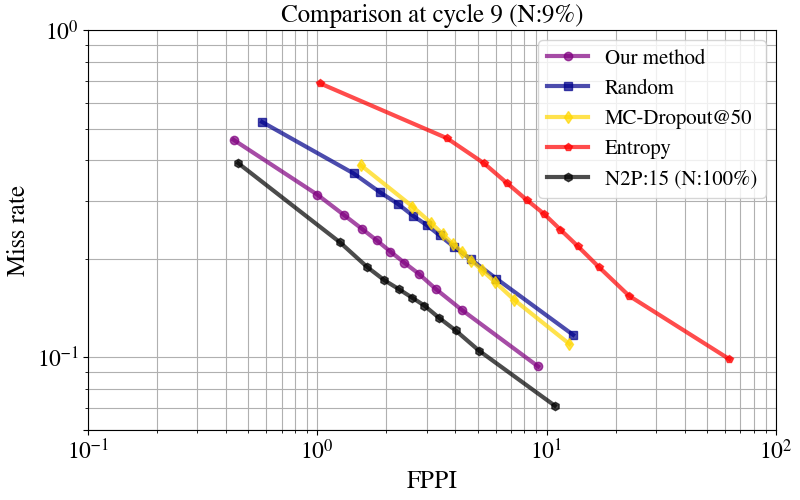}\\
    \includegraphics[width=0.32\linewidth]{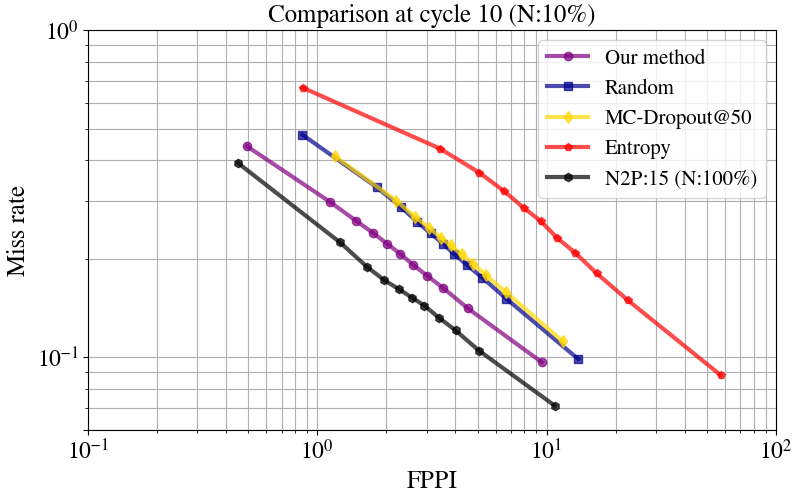}
    \includegraphics[width=0.32\linewidth]{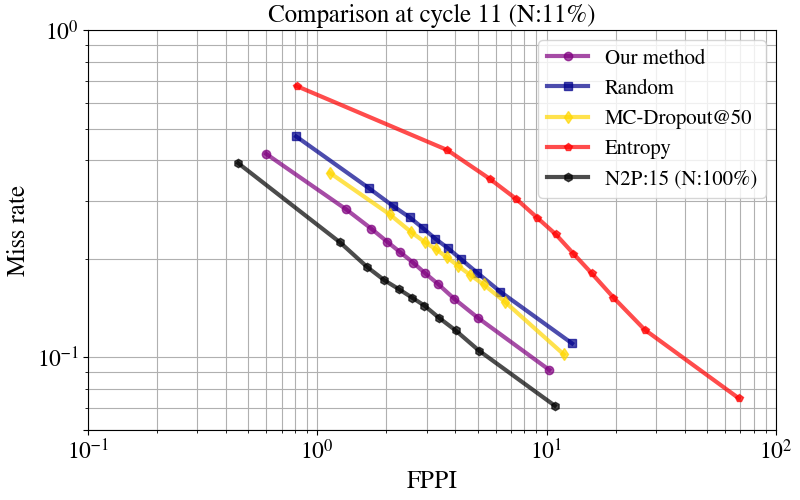}
    \includegraphics[width=0.32\linewidth]{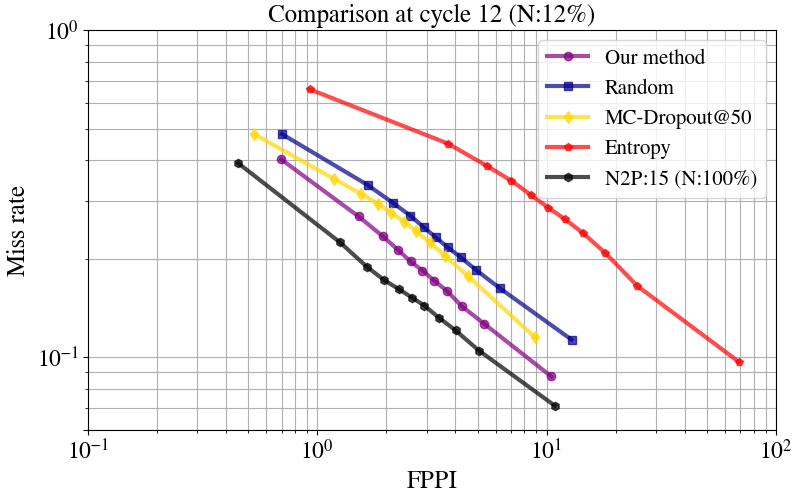}\\
    \includegraphics[width=0.32\linewidth]{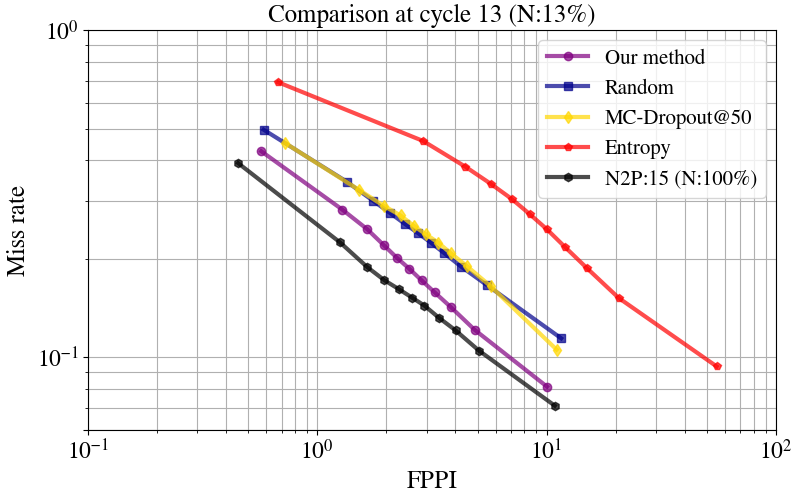}
    \includegraphics[width=0.32\linewidth]{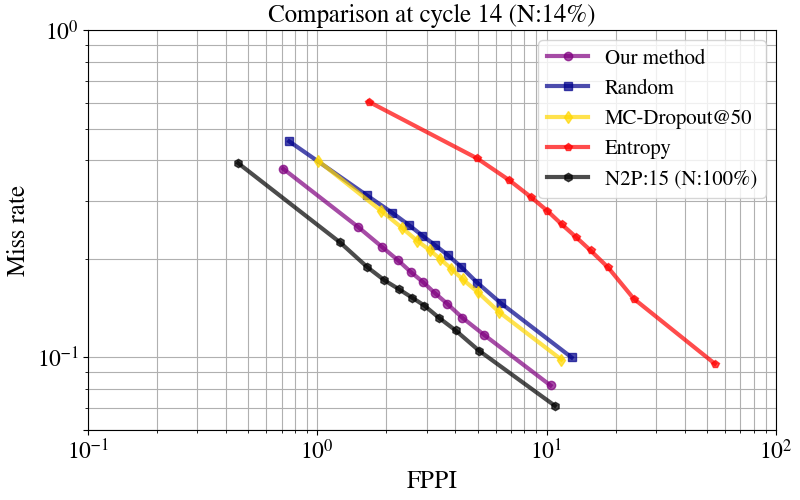}
    \caption{Comparing our method and its MC-Dropout and Entropy variants with the guided random selection at each cycle, for Caltech Pedestrian dataset.}
    \label{fig:caltech_per_cycle_comp}
\end{figure*}

\paragraph{Per cycle comparison.}
In the main submission, we compared our method and its variants of MC-Dropout and binary entropy with the guided random selection at specific cycles. 
In Figure \ref{fig:caltech_per_cycle_comp}, we compared these method at all cycles. We see how our active learning method and the guided random one select images which give rise to detectors of similar accuracy at 1st cycle. However, starting from the 2nd cycle, our method selects images which turn out in a more accurate detector.

\paragraph{Statistics of $\mathcal{X}_{al}$.}
We showed in our experiments that the number of pedestrian instances selected by our method is higher than for the guided random method. At each cycle, we also computed the number of frames in $\mathcal{X}_{al}$ that contains at least one pedestrian instance, both for our active learning method and guided random. Figure \ref{fig:caltech_stat_frames} shows the results.

At the end of cycle 14th, 2895 out of 7K frames ($41\%$ of frames) have at least one pedestrian instance when $\mathcal{X}_{al}$ is selected using our method. In contrast, 1460 out of 7K frames ($21\%$) contain pedestrian instances using the guided random method. 


\section{BDD100K dataset}

\begin{figure}[t!]
		\centering
		\includegraphics[width=0.99\linewidth]{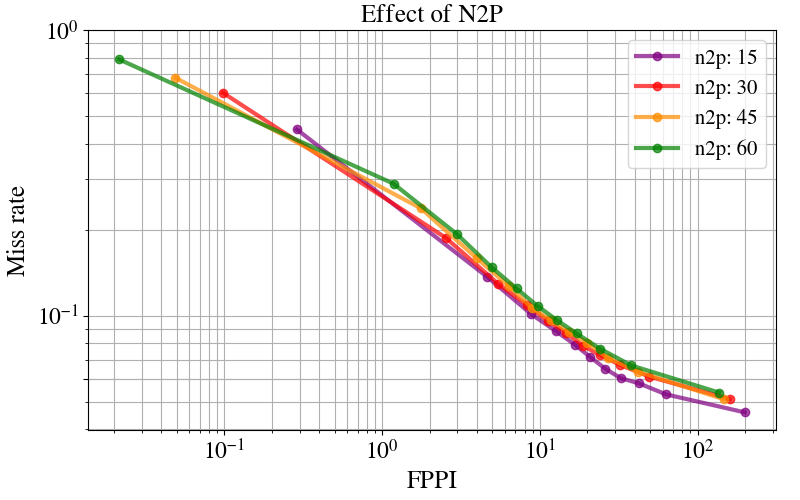}
		\caption{Pedestrian detection performance of our base neural network on the BDD100K dataset, using different N2Ps (negative to positive ratios).}
		\label{fig:exp_bdd_n2p}
	\end{figure}

\paragraph{Lower-bound error.}
Figure \ref{fig:exp_bdd_n2p} illustrates the performance of our network on the BDD100K dataset using different N2Ps. The results show that our method is less accurate on the BDD100K dataset compared to the Caltech Pedestrian dataset. We think this is due to the fact that our network is too lightweight for BDD100K complexity. Thus, our immediate future work is to use a network with higher capacity for this case. 

\begin{figure*}
    \centering
    \includegraphics[width=0.32\linewidth]{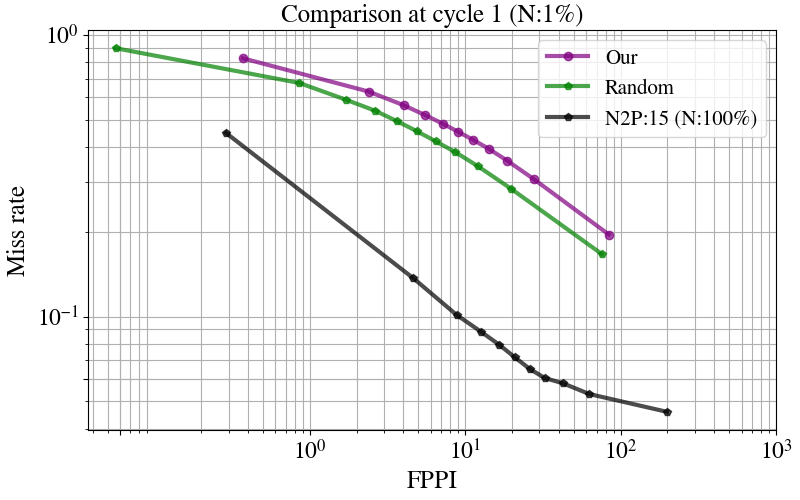}
    \includegraphics[width=0.32\linewidth]{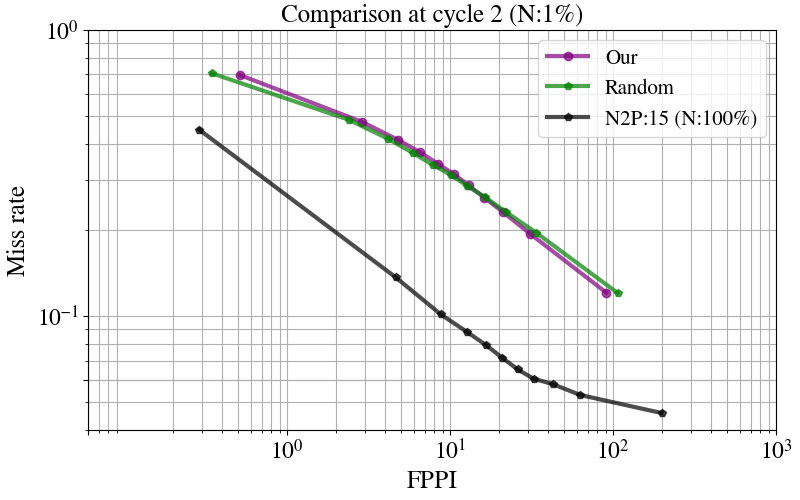}
    \includegraphics[width=0.32\linewidth]{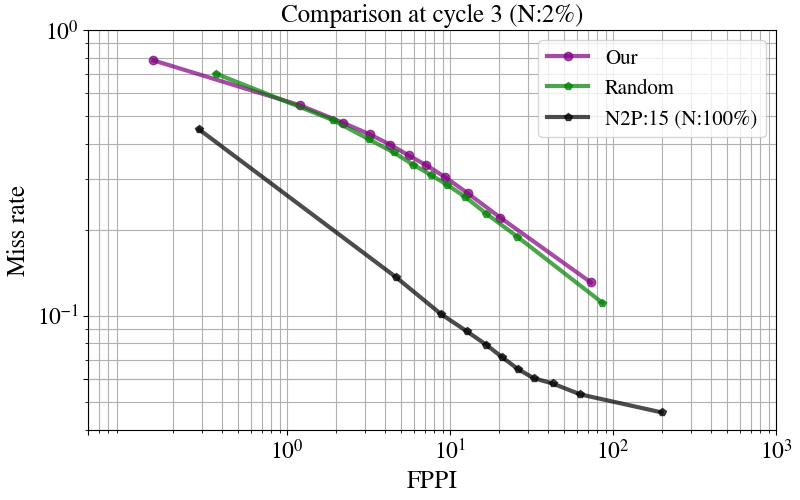}\\
    \includegraphics[width=0.32\linewidth]{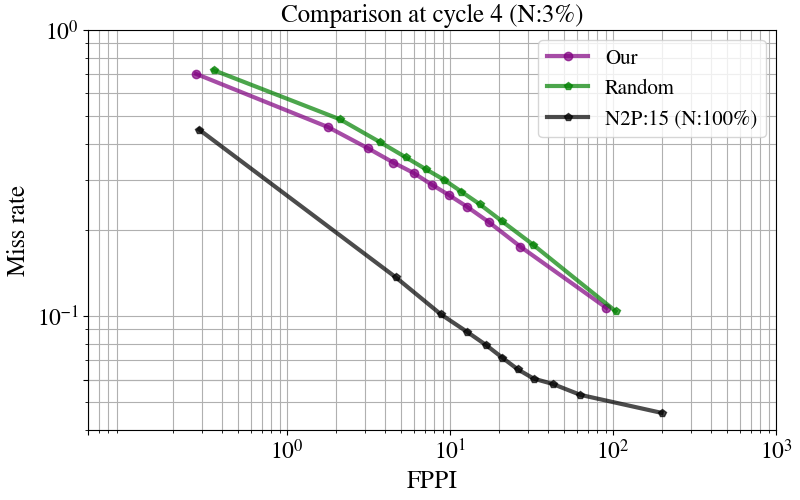}
    \includegraphics[width=0.32\linewidth]{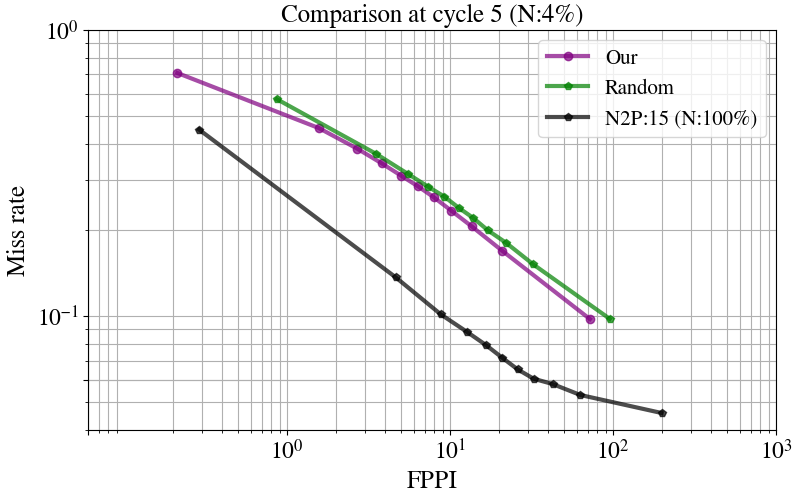}
    \includegraphics[width=0.32\linewidth]{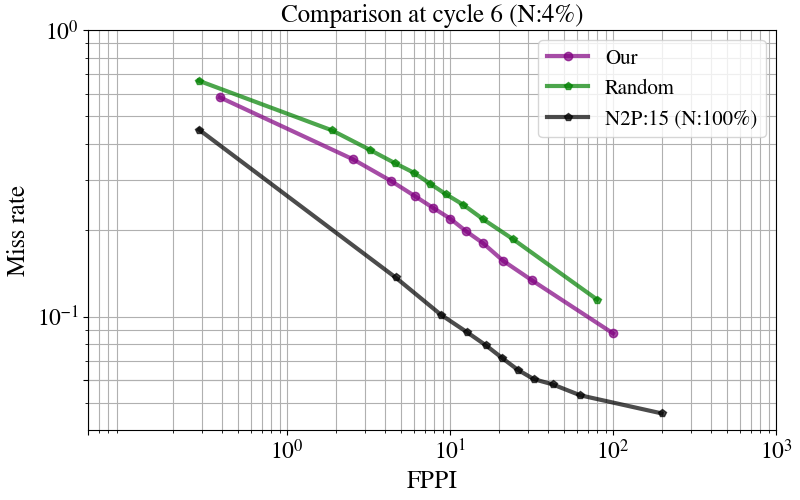}\\
    \includegraphics[width=0.32\linewidth]{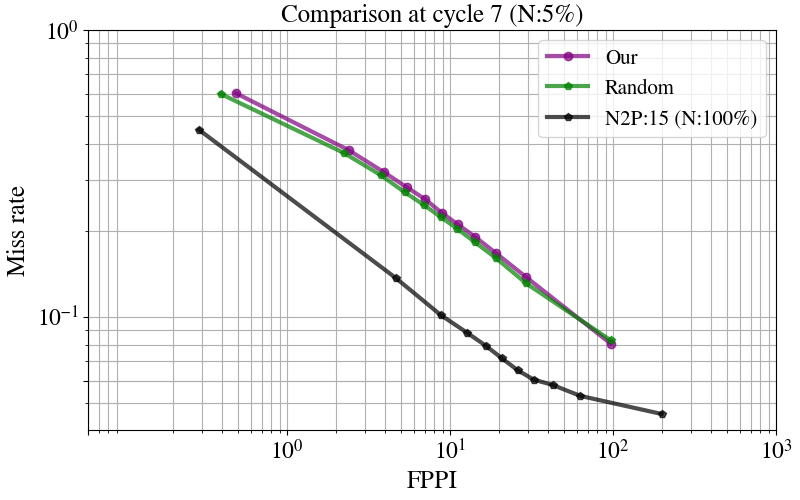}
    \includegraphics[width=0.32\linewidth]{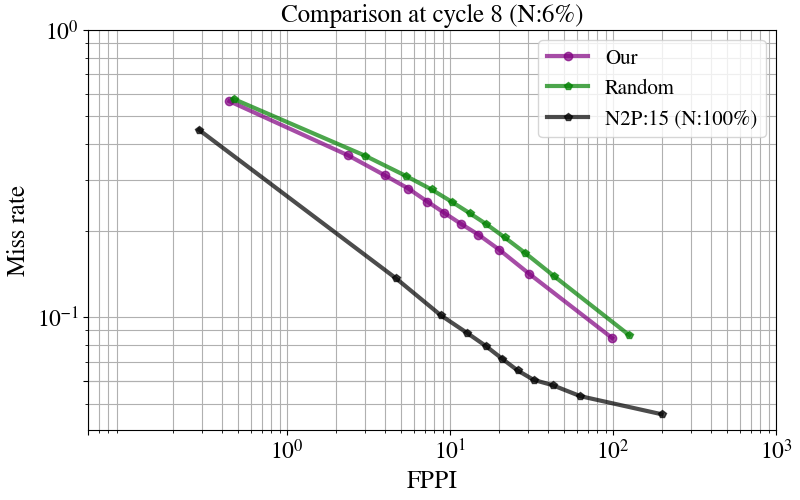}
    \includegraphics[width=0.32\linewidth]{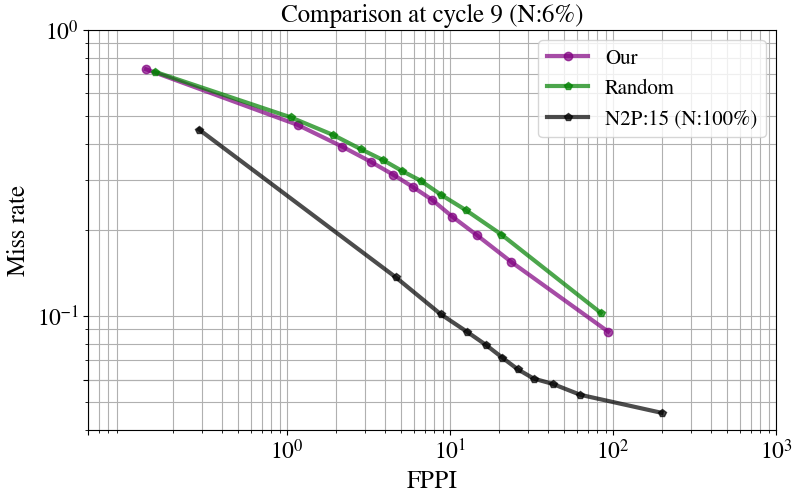}\\
    \includegraphics[width=0.32\linewidth]{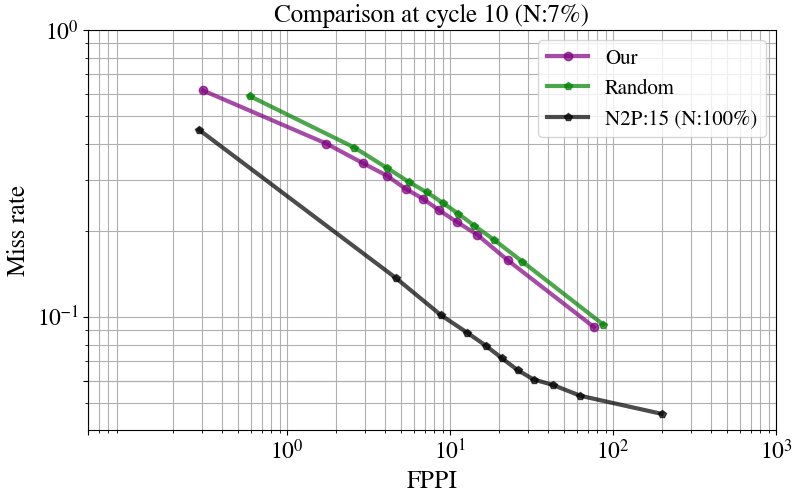}
    \includegraphics[width=0.32\linewidth]{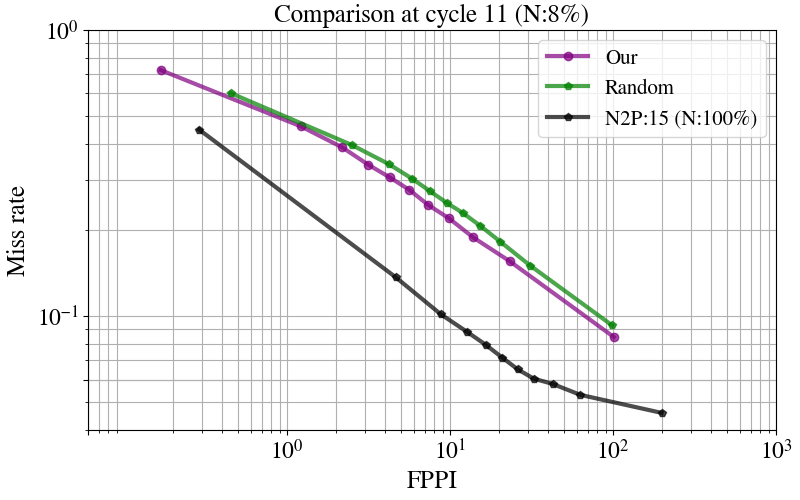}
    \includegraphics[width=0.32\linewidth]{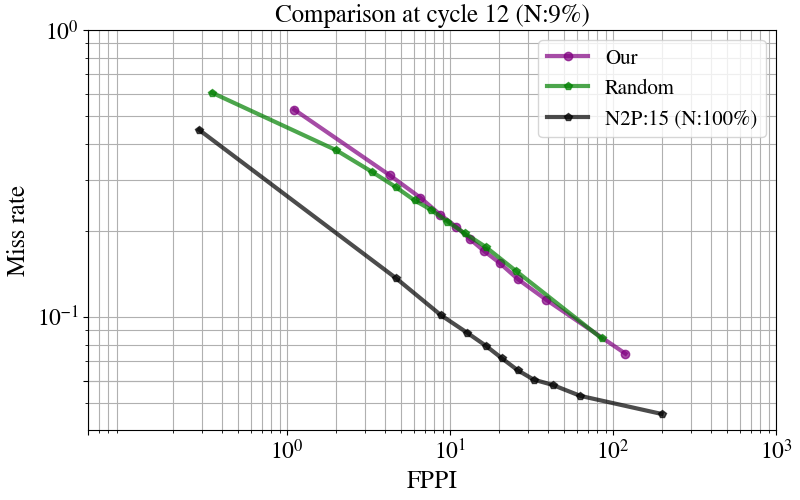}\\
    \includegraphics[width=0.32\linewidth]{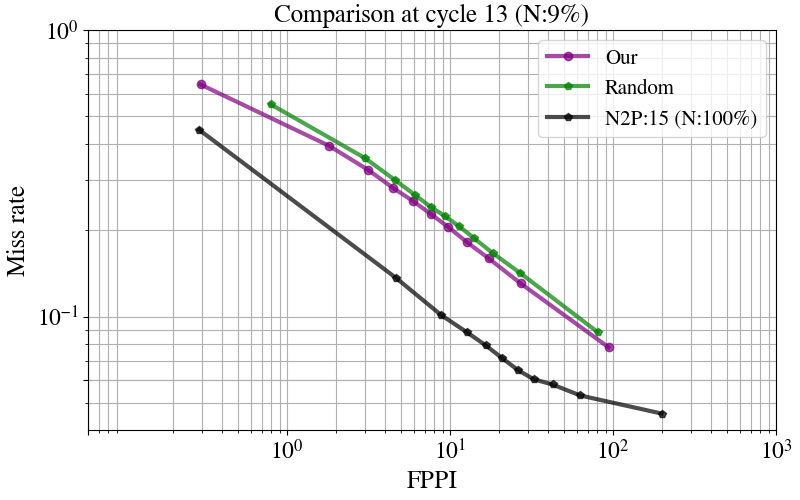}
    \caption{Comparing our method with random selection at each cycle, for BDD100K.}
    \label{fig:bdd_per_cycle_comp}
\end{figure*}

\paragraph{Per cycle comparison.} For BDD100K,
Figure \ref{fig:bdd_per_cycle_comp} compares the detection performance based on the images selected by our active learning method {\vs} the ones selected by the guided random method, at each cycle. The results indicate that our method performs slightly better than the random selection. However, the improvement is not as significant as for the Caltech Pedestrian dataset. We think this is because for this dataset it is required a more complex network architecture able to reduce the bias.

\begin{figure}[t!]
    \centering
    \includegraphics[width=0.99\linewidth]{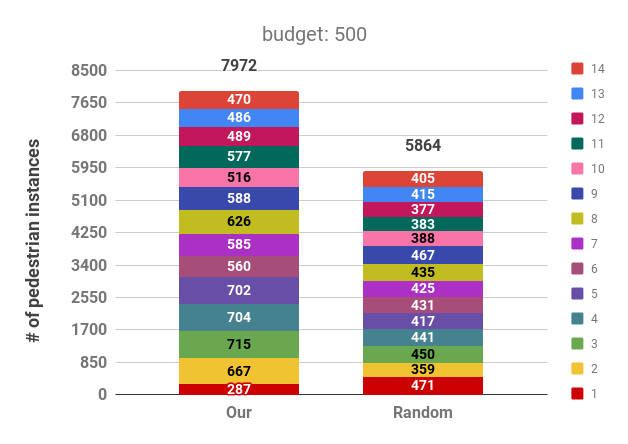}\\
    \includegraphics[width=0.99\linewidth]{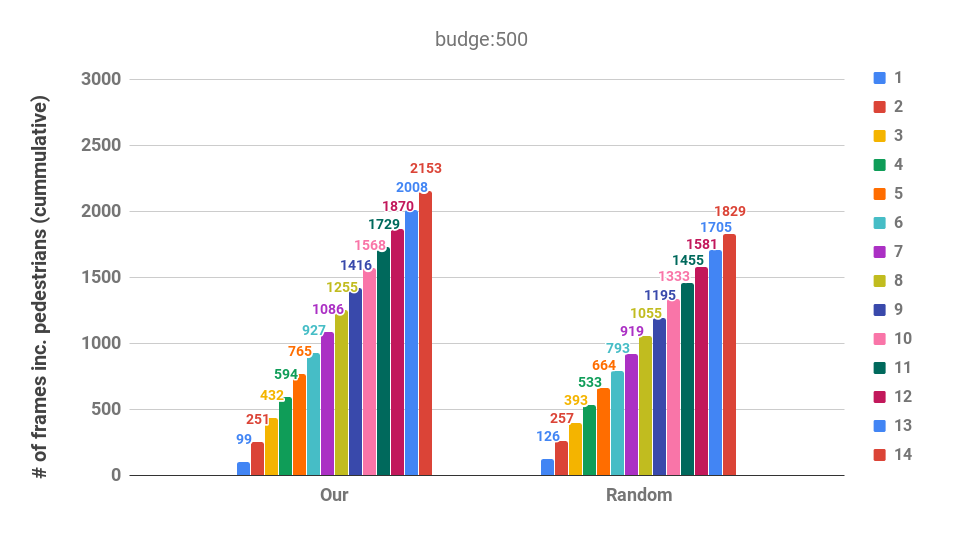}
    \caption{Statistics of $\mathcal{X}_{al}$ at each cycle in terms of number of pedestrian instances (top) and  number of images containing at least one pedestrian instance (bottom) when $\mathcal{X}_u$ is the BDD100K dataset. Bars correspond to cycles.}
    \label{fig:bdd_stats}
\end{figure}

\paragraph{Statistics of $\mathcal{X}_{al}$.}
We also computed the statistics of $\mathcal{X}_{al}$ for each cycle on the BDD100K dataset. Figure \ref{fig:bdd_stats} illustrates the results. Similar to the Caltech Pedestrian datasets, our method selects frames with more pedestrian instances compared to the random selection. Moreover, the number of frames containing  at least one pedestrian instance is higher using our method. 

\bibliographystyle{ieee}
\bibliography{all.bib}